%% file: arxiv.tex
\documentclass[11pt]{article}

\usepackage[margin=1in]{geometry}
\usepackage{times}
\usepackage[hyphens]{url}
\usepackage{graphicx}
\usepackage{natbib}
\usepackage{caption}
\usepackage{hyperref}

\usepackage{amsmath,amssymb,amsfonts}

\usepackage{booktabs}
\usepackage{subcaption}

\usepackage{algorithm}
\usepackage{algorithmic}

\usepackage{xcolor}
\usepackage{cleveref}

\setcitestyle{authoryear,open={(},close={)},citesep={;},aysep={,},yysep={,}}
\renewcommand{\cite}{\citep}

\hypersetup{
    colorlinks=true,
    linkcolor=blue,
    filecolor=magenta,
    urlcolor=cyan,
    citecolor=blue,
}

\title{CAST: Compositional Analysis via Spectral Tracking for Understanding Transformer Layer Functions}

\author{
    Zihao Fu$^{1}$ \quad Ming Liao$^{2}$ \quad Chris Russell$^{3}$ \quad Zhenguang G. Cai$^{1}$ \\
    $^{1}$The Chinese University of Hong Kong \\
    $^{2}$Hong Kong Polytechnic University \\
    $^{3}$University of Oxford \\
    \texttt{zihaofu@cuhk.edu.hk, mliao@polyu.edu.hk,} \\
    \texttt{chris.russell@oii.ox.ac.uk, zhenguangcai@cuhk.edu.hk}
}

\date{}

\begin{document}

\maketitle

\input{main}

\bibliographystyle{plainnat}
\bibliography{references}

\appendix
\input{appendix}

\end{document}

%% file: main.tex
\begin{abstract}
Large language models have achieved remarkable success but remain largely black boxes with poorly understood internal mechanisms. To address this limitation, many researchers have proposed various interpretability methods including mechanistic analysis, probing classifiers, and activation visualization, each providing valuable insights from different perspectives. Building upon this rich landscape of complementary approaches, we introduce CAST (Compositional Analysis via Spectral Tracking), a probe-free framework that contributes a novel perspective by analyzing transformer layer functions through direct transformation matrix estimation and comprehensive spectral analysis. CAST offers complementary insights to existing methods by estimating the realized transformation matrices for each layer using Moore-Penrose pseudoinverse and applying spectral analysis with six interpretable metrics characterizing layer behavior. Our analysis reveals distinct behaviors between encoder-only and decoder-only models, with decoder models exhibiting compression-expansion cycles while encoder models maintain consistent high-rank processing. Kernel analysis further demonstrates functional relationship patterns between layers, with CKA similarity matrices clearly partitioning layers into three phases: feature extraction, compression, and specialization.
\end{abstract}

\section{Introduction}

Large language models have achieved remarkable success across diverse tasks~\citep{radford2019language,liu2019roberta,touvron2023llama}, yet their internal mechanisms remain poorly understood~\citep{rogers2020primer}. Critical questions about how transformer layers process information, their computational roles, and information flow continue to challenge researchers~\citep{tenney2019bert,kovaleva2019revealing}.

\begin{figure}[t]
\centering
\includegraphics[width=0.65\textwidth]{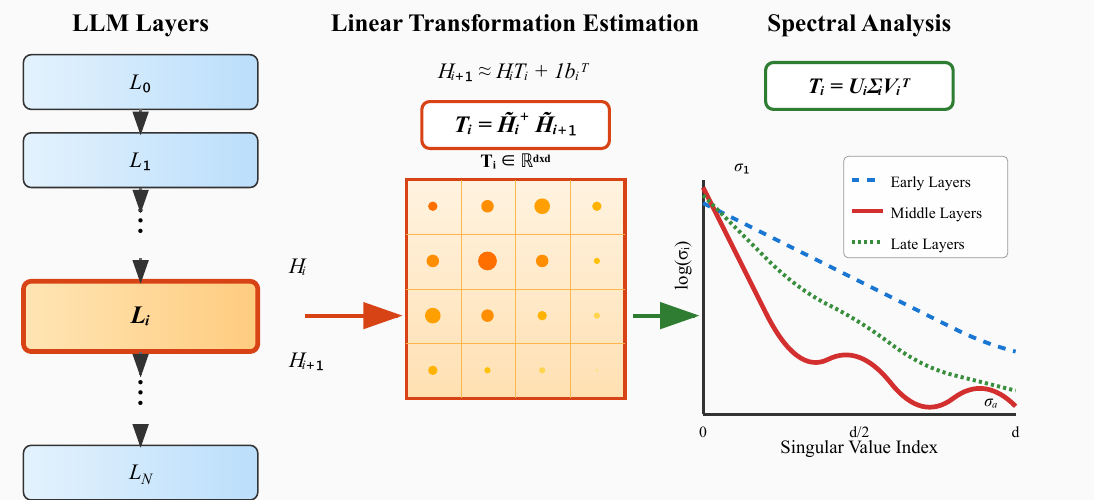}
\caption{Overview of the CAST framework. Given a large language model with multiple layers, we focus on layer $L_i$ and estimate the linear transformation between its input $H_i$ and output $H_{i+1}$ representations. Spectral analysis reveals patterns across layer types: early layers maintain high effective rank, middle layers perform aggressive compression, late layers show moderate specialization.}
\label{fig:cast_overview}
\vspace{-1em}
\end{figure}

To make LLMs more interpretable, many researchers have proposed various methods to understand the internal mechanisms of these models, each contributing valuable perspectives to our understanding. The logit lens~\citep{nostalgebraist2020logit} projects intermediate hidden states to vocabulary space to trace prediction evolution, while the tuned lens~\citep{belrose2023eliciting} learns affine transformations for better alignment between layers. Probing classifiers~\citep{belinkov2019analysis,hewitt2019structural} decode linguistic properties from representations, while attention visualization tools~\citep{vig2019multiscale,vig2019analyzing} provide insights into attention mechanisms. These approaches have significantly advanced our understanding of transformer behavior, yet they primarily focus on specific aspects: probe-dependent approaches examine static representational properties, projection methods analyze output-oriented behavior, and visualization techniques illuminate attention patterns. This diversity of perspectives highlights the complexity of transformer interpretation and suggests that comprehensive understanding requires multiple complementary analytical lenses.

To further strengthen the understanding of transformer internals through transformation-centric analysis, we propose CAST (Compositional Analysis via Spectral Tracking), an analysis framework that examines transformer layer dynamics through direct transformation matrix estimation and spectral decomposition~\citep{ethayarajh2019how,voita2019bottom}. As illustrated in Figure~\ref{fig:cast_overview}, rather than analyzing what information is encoded in representations as in probe-based approaches~\citep{belinkov2019analysis,hewitt2019structural} or how it relates to outputs as in projection methods~\citep{nostalgebraist2020logit,belrose2023eliciting}, CAST examines how transformer layers actively transform their inputs, providing insights that complement existing approaches~\citep{zhou2021directprobe}. Though transformer layers are highly non-linear, CAST uses linear estimation because linear transformation constitutes the major component of layer processing~\citep{elhage2021mathematical,olah2020zoom}, as we demonstrate through residual norm analysis showing that linear approximation captures substantial transformation behavior. The framework contains two key components: the first is linear transformation estimation that uses Moore-Penrose pseudoinverse~\citep{golub2013matrix} to directly estimate transformation matrices between consecutive layers, and the second is comprehensive spectral analysis~\citep{denton2014exploiting,bloom2023singular} to extract six designed metrics capturing spectral distributions and transformation characteristics from a transformation-centric viewpoint.

We conduct extensive experiments on four representative transformer architectures including GPT-2~\citep{radford2019language}, RoBERTa~\citep{liu2019roberta}, Llama~\citep{touvron2023llama}, and DeepSeek-R1~\citep{deepseek2025}. Interestingly, we find that decoder-only models (GPT-2, Llama, DeepSeek-R1) exhibit consistent compression-expansion cycles through their layers, with effective rank dropping sharply at middle layers before recovering, consistent with information processing theories~\citep{tishby2015deep,schwartz2017opening}, while encoder-only models (RoBERTa) maintain high effective rank throughout their depth. This architectural distinction reveals fundamentally different information processing strategies: decoders implement an information bottleneck for next-token prediction~\citep{voita2019bottom,tenney2019bert}, while encoders preserve comprehensive representations for downstream tasks~\citep{rogers2020primer,kovaleva2019revealing}. Moreover, our kernel analysis demonstrates that middle compression layers in decoders exhibit the strongest nonlinearity, suggesting complex transformations during abstraction.

Our contributions are: (1) We propose CAST, a probe-free framework examining transformation dynamics between layers through direct matrix estimation and spectral analysis, complementing existing static or output-focused methods. (2) We discover distinct architectural patterns: decoders exhibit compression-expansion cycles with three functional phases, while encoders maintain consistent high-rank processing. (3) We provide validation across GPT-2, RoBERTa, Llama, and DeepSeek-R1, revealing how transformation properties reflect architectural objectives.

\section{Related Work} 

\begin{table}[t]
\centering
\scriptsize
\caption{Comparison of transformer interpretability approaches. CAST provides complementary transformation-centric analysis.}
\label{tab:method_comparison}
\begin{tabular}{@{}lccccc@{}}
\toprule
& \textbf{Probe} & \textbf{Decomp.} & \textbf{Direct} & \textbf{Mech.} & \textbf{CAST} \\
\midrule
Analyzes representations & \textcolor{green}{\checkmark} & -- & \textcolor{green}{\checkmark} & \textcolor{green}{\checkmark} & -- \\
Analyzes transformations & -- & -- & -- & -- & \textcolor{green}{\checkmark} \\
Fine-grained analysis & \textcolor{green}{\checkmark} & -- & -- & \textcolor{green}{\checkmark} & -- \\
Semantic outputs & \textcolor{green}{\checkmark} & -- & -- & \textcolor{green}{\checkmark} & -- \\
Requires training & \textcolor{green}{\checkmark} & -- & -- & \textcolor{green}{\checkmark} & -- \\
Task-independent & -- & \textcolor{green}{\checkmark} & \textcolor{green}{\checkmark} & -- & \textcolor{green}{\checkmark} \\
Quantitative metrics & \textcolor{green}{\checkmark} & \textcolor{green}{\checkmark} & \textcolor{green}{\checkmark} & -- & \textcolor{green}{\checkmark} \\
Cross-layer patterns & -- & -- & \textcolor{green}{\checkmark} & -- & \textcolor{green}{\checkmark} \\
Distinguishes arch. & -- & -- & -- & -- & \textcolor{green}{\checkmark} \\
\bottomrule
\end{tabular}
\vspace{-3mm}
\end{table}

\textbf{Probe-Based Analysis.} These methods analyze representations through auxiliary models that require training. The logit lens~\citep{nostalgebraist2020logit} projects hidden states to vocabulary space for semantic outputs, while the tuned lens~\citep{belrose2023eliciting} learns affine transformations for better layer alignment. Probing classifiers~\citep{belinkov2019analysis,hewitt2019structural} provide fine-grained analysis of linguistic properties, with \citet{tenney2019bert} showing BERT rediscovers the classical NLP pipeline. While these approaches offer quantitative metrics, they are task-dependent—probes may learn superficial patterns~\citep{belinkov2022probing}, high accuracy doesn't guarantee task relevance~\citep{ravichander2021probing}, and amnesic probing shows encoded information isn't necessarily used~\citep{elazar2021amnesic}. Recent work like Patchscopes~\citep{ghandeharioun2024patchscopes} uses activation patching. Unlike these representation-focused methods, CAST analyzes transformation dynamics without requiring training.

\textbf{Matrix Decomposition in Neural Networks.} These task-independent methods provide quantitative metrics without requiring training. \citet{denton2014exploiting} applied SVD for network compression, achieving significant speedup. Recent approaches include Joint SVD~\citep{chen2022joint} and AdaSVD~\citep{adasvd2025} for adaptive compression. \citet{bloom2023singular} showed SVD of transformer weights yields interpretable singular vectors. Spectral analysis connects eigenvalue patterns to network behavior~\citep{johansson2022exact}, while intrinsic dimension analysis~\citep{ansuini2019intrinsic} reveals non-linear evolution through layers. However, these methods neither analyze representations nor transformations, instead focusing on specific parameter weights. CAST extends this approach by applying SVD to estimated transformation matrices during forward passes.

\textbf{Direct Analysis Methods.} These task-independent approaches analyze representations without training requirements. DirectProbe~\citep{zhou2021directprobe} examines representation geometry directly, providing quantitative metrics. Geometric approaches measure anisotropy~\citep{ethayarajh2019how}, while information-theoretic methods~\citep{voita2019bottom} reveal cross-layer patterns in representation evolution. \citet{jiang2020can} applied information bottleneck theory for attribution. Recent work~\citep{razzhigaev2024shape} reveals patterns through direct SVD, while \citet{machina2024anisotropy} challenges anisotropy assumptions. Surveys~\citep{rogers2020primer} consolidate probe-free approaches including similarity methods (CKA, CCA, Procrustes) that capture cross-layer relationships. While these methods analyze static representational properties, CAST focuses on transformation dynamics and distinguishes architectural behaviors.

\textbf{Mechanistic Interpretability.} These methods provide fine-grained analysis at neuron level with semantic outputs, though requiring training. \citet{conmy2023automated} introduce automated circuit discovery, producing interpretable semantic outputs like GPT-2's greater-than circuit. Sparse autoencoders~\citep{cunningham2023sparse,bricken2023monosemanticity} decompose activations into interpretable features through learned decompositions. \citet{templeton2024scaling} scale these approaches to production models, while \citet{olah2023interpretability} envision transformers as interpretable circuits. These methods analyze representations at microscopic scale but are task-dependent. Unlike mechanistic approaches that focus on neuron-level semantic outputs, CAST provides macroscopic transformation analysis without training requirements.

\section{Method}
CAST (Compositional Analysis via Spectral Tracking) provides a probe-free framework for understanding transformer layer functions through direct transformation matrix estimation and spectral analysis. Although transformer layers exhibit complex non-linear behaviors, we employ linear approximation as the linear component almost constitutes the dominant transformation mechanism, as validated by our residual analysis in \Cref{sec:layer_characterization}

The framework consists of three core components: \textbf{Linear Transformation Estimation} using Moore-Penrose pseudoinverse to directly estimate layer-to-layer transformation matrices from hidden states; \textbf{Spectral Analysis} applying spectral methods to extract six interpretable metrics characterizing transformation properties; and \textbf{Kernel Analysis} examining non-linear aspects through complementary kernel methods to validate linear approximations and reveal transformation complexity patterns.

\subsection{Linear Transformation Approximation}

Given a large language model with layers $L_0, L_1, \ldots, L_N$ where $L_0$ represents the input embedding layer and $L_N$ the final output layer, we consider an input sequence $\mathbf{x} = (x_1, x_2, \ldots, x_s) \in \mathbb{R}^s$ of length $s$ tokens. After processing through layer $L_i$ (where $i$ denotes the layer index), we obtain hidden representations $P_i^{(k)} \in \mathbb{R}^{s \times d}$ for the $k$-th sequence, where $d$ is the model's hidden dimension. Each row $P_i^{(k)}[j, :] \in \mathbb{R}^d$ represents the $d$-dimensional hidden state for token $j$ at layer $i$ in sequence $k$.

For a dataset of $n$ input sequences, we concatenate the hidden states from all sequences by stacking them vertically: $H_i = [P_i^{(1)}; P_i^{(2)}; \ldots; P_i^{(n)}] \in \mathbb{R}^{m \times d}$ where $m = n \times s$ represents the total number of token representations across all sequences. This concatenated matrix $H_i$ contains all hidden states at layer $i$, with rows indexed from 1 to $m$. We model transformation from layer $i$ to layer $i+1$ as affine transformation, where output representations are approximated by linear mapping of input representations plus bias:

\begin{equation}
H_{i+1} \approx H_i T_i + \mathbf{1} b_i^T
\end{equation}

where $H_i \in \mathbb{R}^{m \times d}$ contains the hidden states from layer $i$, $T_i \in \mathbb{R}^{d \times d}$ is the transformation matrix we seek to estimate, $b_i \in \mathbb{R}^d$ is the bias vector, and $\mathbf{1} \in \mathbb{R}^m$ is a vector of ones. To isolate the linear component, we center both input and output representations by subtracting their respective column-wise means, effectively removing the bias term:

\begin{equation}
\widetilde{H}_i = H_i - \mathbf{1} \mu_i^T, \quad \widetilde{H}_{i+1} = H_{i+1} - \mathbf{1} \mu_{i+1}^T
\end{equation}

where $\mu_i = \frac{1}{m} \sum_{j=1}^m H_i[j, :] \in \mathbb{R}^d$ is the mean hidden state at layer $i$. We then estimate the transformation matrix using the Moore-Penrose pseudoinverse:

\begin{equation}
T_i = \widetilde{H}_i^{\dagger} \widetilde{H}_{i+1}
\end{equation}

where $(\cdot)^{\dagger}$ represents the pseudoinverse operation. This choice of estimator minimizes the Frobenius norm of the reconstruction error $\|\widetilde{H}_{i+1} - \widetilde{H}_i T_i\|_F$ while handling potential rank deficiency in the hidden state matrices. The resulting transformation matrix $T_i$ captures the linear component of how layer $i$ transforms input to produce representations observed at layer $i+1$, providing mathematical characterization of computational operations performed by each layer.

\subsection{Spectral Analysis}
\label{sec:spectral_analysis}

Once we have estimated the transformation matrix $T_i$ for each layer, we apply singular value decomposition to reveal its spectral structure:

\begin{equation}
T_i = U_i \Sigma_i V_i^T
\end{equation}

where $U_i, V_i \in \mathbb{R}^{d \times d}$ are orthogonal matrices representing the left and right singular vectors respectively, and $\Sigma_i \in \mathbb{R}^{d \times d}$ is a diagonal matrix containing the singular values $\sigma_1 \geq \sigma_2 \geq \ldots \geq \sigma_d \geq 0$. The singular values quantify the strength of each transformation direction, while the singular vectors define the principal axes along which the transformation operates. From this decomposition, we extract six key metrics that characterize the transformation properties:

\textbf{Effective Rank (ER)} counts singular values exceeding a threshold relative to the maximum singular value to measure the intrinsic dimensionality of the transformation~\citep{roy2007effective}. The effective rank serves as a real-valued extension of matrix rank with roots in information theory, representing the true effective dimension of the vector space by accounting for the relative importance of different dimensions through their eigenvalue magnitudes~\citep{udell2019matrix}. A high effective rank indicates that the layer spreads information across many dimensions (feature expansion), while a low effective rank suggests dimensional compression where the transformation projects inputs into a lower-dimensional subspace~\citep{golub2013matrix,udell2019matrix}. This metric reveals whether layers expand features for richer representation or compress for abstraction.

\textbf{Spectral Decay Rate (SDR)} fits $\log(\sigma_i) = -\alpha i + \beta$ to quantify how rapidly singular values decrease with rank. The decay rate of singular values relates to the complexity and ill-posedness of matrix transformations, with faster decay indicating more regular and structured transformations~\citep{drineas2016fast,ubaru2016fast}. A steep decay (high $\alpha$) indicates aggressive compression where only a few dominant directions are preserved, while a gentle decay suggests more uniform utilization of transformation directions. This captures the compression strategy employed by each layer—whether it performs sharp bottlenecking or gradual dimensionality reduction.

\textbf{Transformation Entropy (TE)} computes $H = -\sum_i p_i \log p_i$ where $p_i = \sigma_i/\sum_j \sigma_j$ to assess the distributional complexity of singular values. High entropy indicates that transformation strength is spread relatively evenly across many directions (complex, multi-faceted processing), while low entropy suggests concentration in few dominant directions (focused, specialized processing), consistent with information-theoretic principles of diversity and specialization~\citep{schwartz2017opening,bengio2013representation}. This reveals whether a layer performs complex multi-directional transformations or simple unidirectional operations.

\textbf{Anisotropy Index (AI)} measures directional bias as $(\sigma_{\max} - \sigma_{\min})/\sigma_{\text{mean}}$ to quantify how unevenly the transformation treats different input directions. High anisotropy indicates strong directional preferences where certain input patterns are amplified much more than others, while low anisotropy suggests more isotropic processing~\citep{ethayarajh2019how,machina2024anisotropy}. This captures whether the layer developed specialized sensitivities to particular input patterns or processes all directions uniformly.

\textbf{Information Concentration (IC)} applies the Gini coefficient to singular values to quantify inequality in their distribution, computed as $G = \frac{2\sum_i i\sigma_i}{n\sum_i \sigma_i} - \frac{n+1}{n}$. High concentration (approaching 1) indicates that most transformation power is concentrated in very few singular values (highly specialized processing), while low concentration (approaching 0) suggests more democratic distribution of transformation strength (generalized processing). This reveals the degree of functional specialization within the layer.

\textbf{Residual Norm (RN)} computes $\|H_{i+1} - T_i H_i\|_F / \|H_{i+1}\|_F$ to measure the proportion of the layer's output that cannot be explained by linear transformation. A high residual norm indicates substantial nonlinear processing that goes beyond simple linear projection, arising from attention mechanisms and activation functions~\citep{elhage2021mathematical}, while a low residual norm suggests the layer's behavior is well-approximated by linear operations. This quantifies the degree of nonlinearity and computational complexity in the layer's transformations.

These metrics together provide a comprehensive characterization of how each layer transforms information, revealing patterns of compression, expansion, specialization, and nonlinearity across the transformer architecture.

\subsection{Kernel Analysis}

While linear transformation analysis captures the primary mode of information processing, transformer layers exhibit rich nonlinear dynamics that require complementary analysis techniques. To address this limitation, CAST incorporates kernel analysis that examines transformation properties through different mathematical lenses, revealing aspects of layer behavior invisible to purely linear methods.

We extend CAST to kernel space using Random Fourier Features (RFF)~\citep{rahimi2007random}, which provides a scalable approximation of kernel methods through explicit feature maps. Given hidden states $H_i \in \mathbb{R}^{m \times d}$ and $H_{i+1} \in \mathbb{R}^{m \times d}$, we generate $D$ random features by sampling weights $\omega_j \in \mathbb{R}^d$ for $j = 1, \ldots, D$. For RBF kernels, $\omega_j \sim \mathcal{N}(0, 2\gamma I_d)$ where $I_d$ is the $d \times d$ identity matrix. For Laplacian kernels, each component is sampled from the Cauchy distribution: $\omega_{jk} = \gamma \tan(\pi(u_{jk} - 0.5))$ where $u_{jk} \sim \text{Uniform}(0, 1)$. The bandwidth parameter $\gamma$ is computed using the median heuristic: $\gamma = 1/(2 \cdot \text{median}(\|x_i - x_j\|)^2)$. The RFF transformation maps inputs to an explicit $D$-dimensional feature space: $z(x) = \sqrt{2/D} [\cos(\omega_1^T x + b_1), \ldots, \cos(\omega_D^T x + b_D)]^T$ where $b_j \sim \text{Uniform}(0, 2\pi)$. This approximates the kernel function as $k(x, y) \approx z(x)^T z(y)$. We then estimate the transformation matrix in RFF space: $T_{RFF} = Z_{out}^T (Z_{in}^{\dagger})^T$ where $Z_{in} = [z(h_1), \ldots, z(h_m)]^T \in \mathbb{R}^{m \times D}$ and $Z_{out} = [z(h'_1), \ldots, z(h'_m)]^T \in \mathbb{R}^{m \times D}$ are the RFF-transformed representations of all hidden states, with $h_i$ and $h'_i$ denoting the $i$-th row of $H_i$ and $H_{i+1}$ respectively. The crucial insight is that we apply identical spectral analysis to $T_{RFF} \in \mathbb{R}^{D \times D}$: computing SVD to extract singular values and deriving the same six metrics. The kernel residual norm $\|Z_{out} - Z_{in}T_{RFF}^T\|_F / \|Z_{out}\|_F$ quantifies how well the kernel transformation captures nonlinear dynamics.

Additionally, we employ Centered Kernel Alignment (CKA)~\citep{kornblith2019similarity} to quantify similarity between different layers. For layers $i$ and $j$, we compute their respective kernel matrices $K_i = k(H_i, H_i)$ and $K_j = k(H_j, H_j)$ from the hidden states at those layers, then calculate: $\text{CKA}(K_i, K_j) = \frac{\text{tr}(\tilde{K}_i \tilde{K}_j)}{\sqrt{\text{tr}(\tilde{K}_i^2) \text{tr}(\tilde{K}_j^2)}}$ where $\text{tr}(\cdot)$ denotes the matrix trace operation and $\tilde{K}_i$, $\tilde{K}_j$ are the centered versions of the kernel matrices. This reveals functional similarity patterns across the architecture. CKA analysis demonstrates that layers within the same functional phase exhibit high similarity, while layers across phase boundaries show distinct patterns, enabling automatic identification of the three-phase architecture (feature extraction, compression, specialization).

\section{Experiments}

\subsection{Experimental Setup}

We conduct experiments on WikiText-103~\citep{merity2017pointer}, a large corpus of verified Wikipedia articles. We randomly sample sequences with appropriate truncation for computational efficiency. We conduct our analysis on GPT-2~\citep{radford2019language}, RoBERTa-base~\citep{liu2019roberta}, Llama-3.2-1B~\citep{touvron2023llama}, and DeepSeek-R1-Distill-Qwen-1.5B~\citep{deepseek2025}. For transformation estimation, we extract hidden states after layer normalization but before residual connections, using batch size 32 to accumulate 2000 sequences for stable pseudoinverse computation. We compute six metrics from the resulting transformation matrices as detailed in \Cref{sec:spectral_analysis}. All experiments use mixed precision computation on NVIDIA A6000 GPUs. We fix random seeds across data sampling, model initialization, and dropout for reproducibility.

\begin{figure*}[t]
\centering
\includegraphics[width=0.19\textwidth]{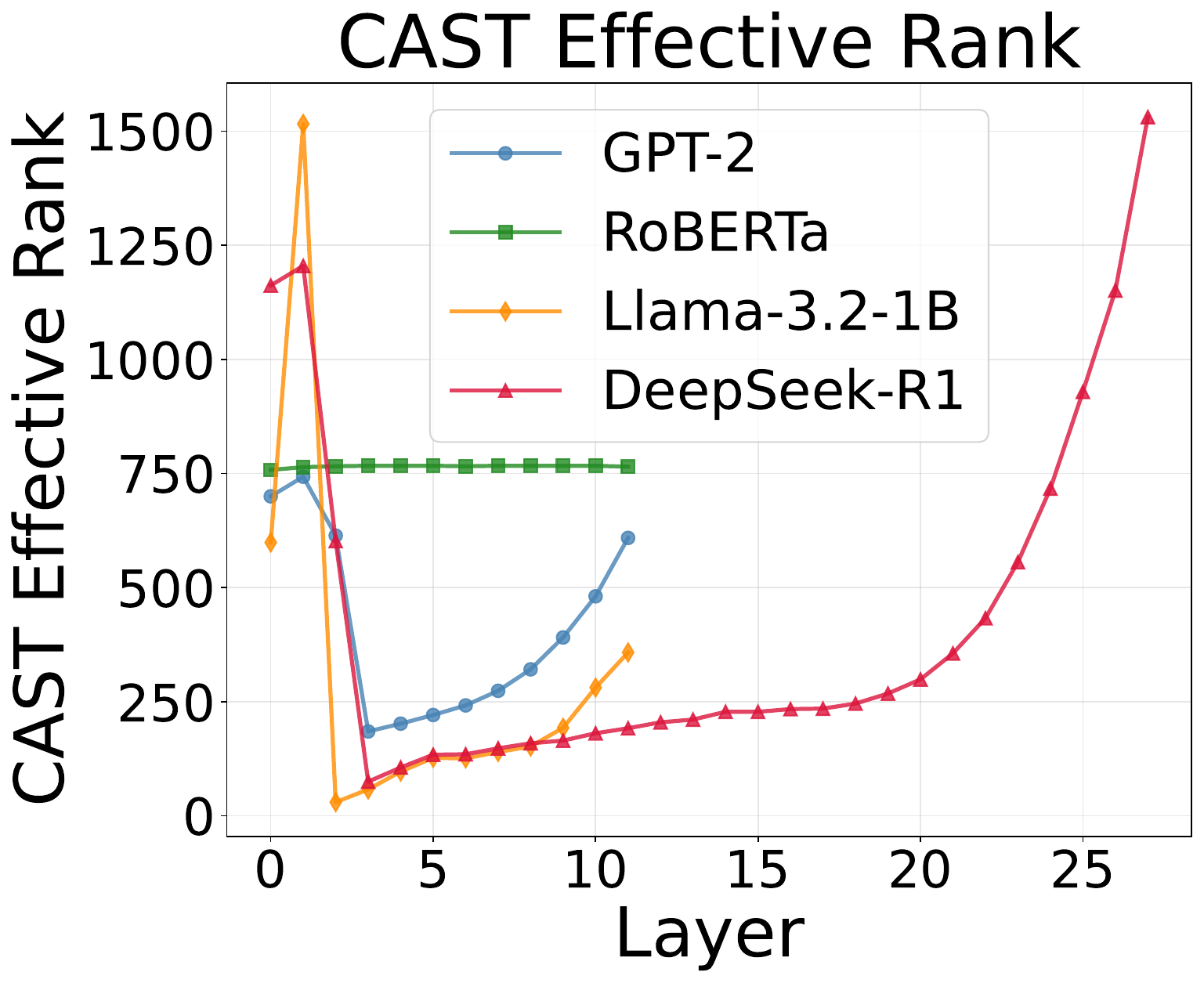}
\hfill
\includegraphics[width=0.19\textwidth]{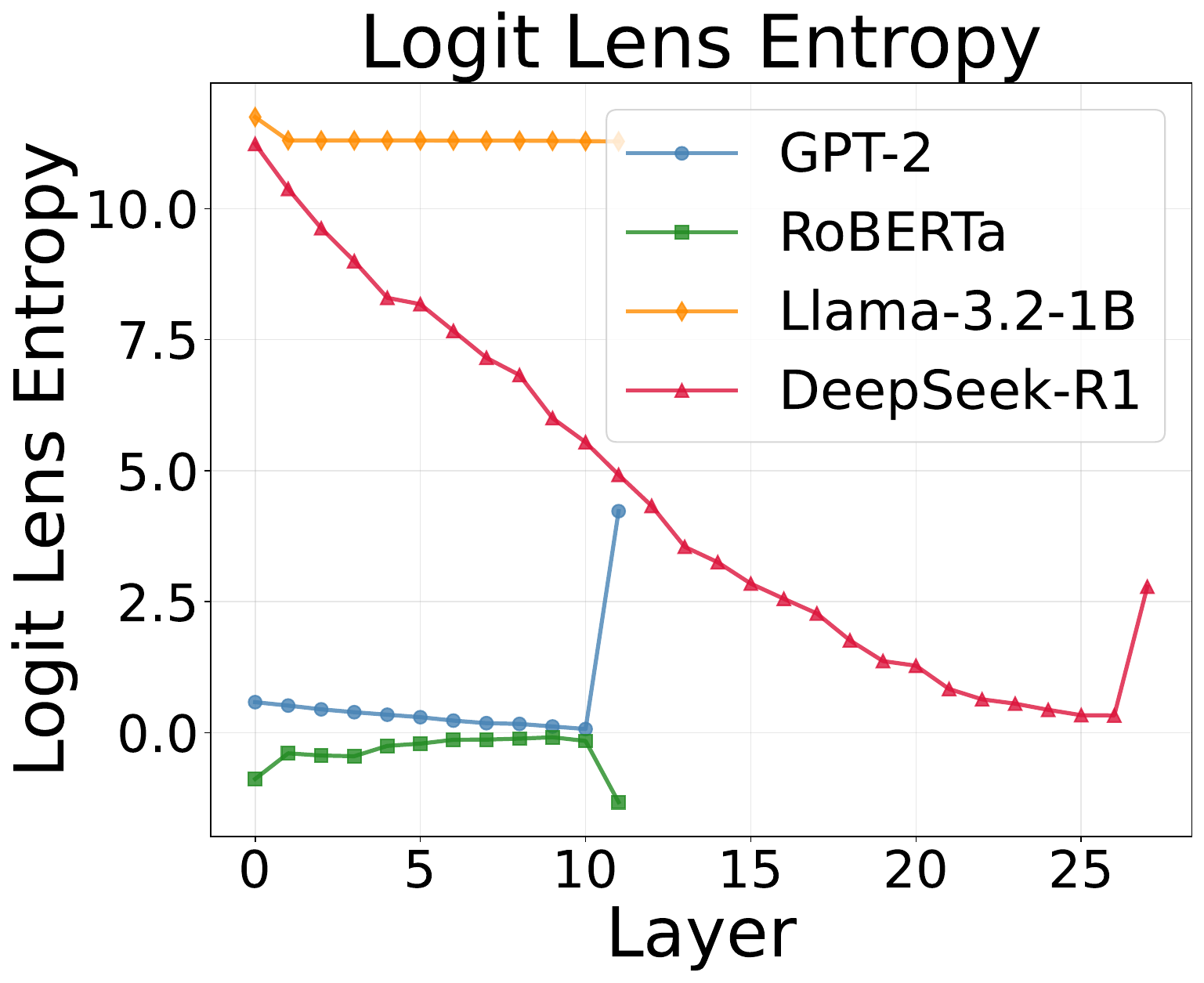}
\hfill
\includegraphics[width=0.19\textwidth]{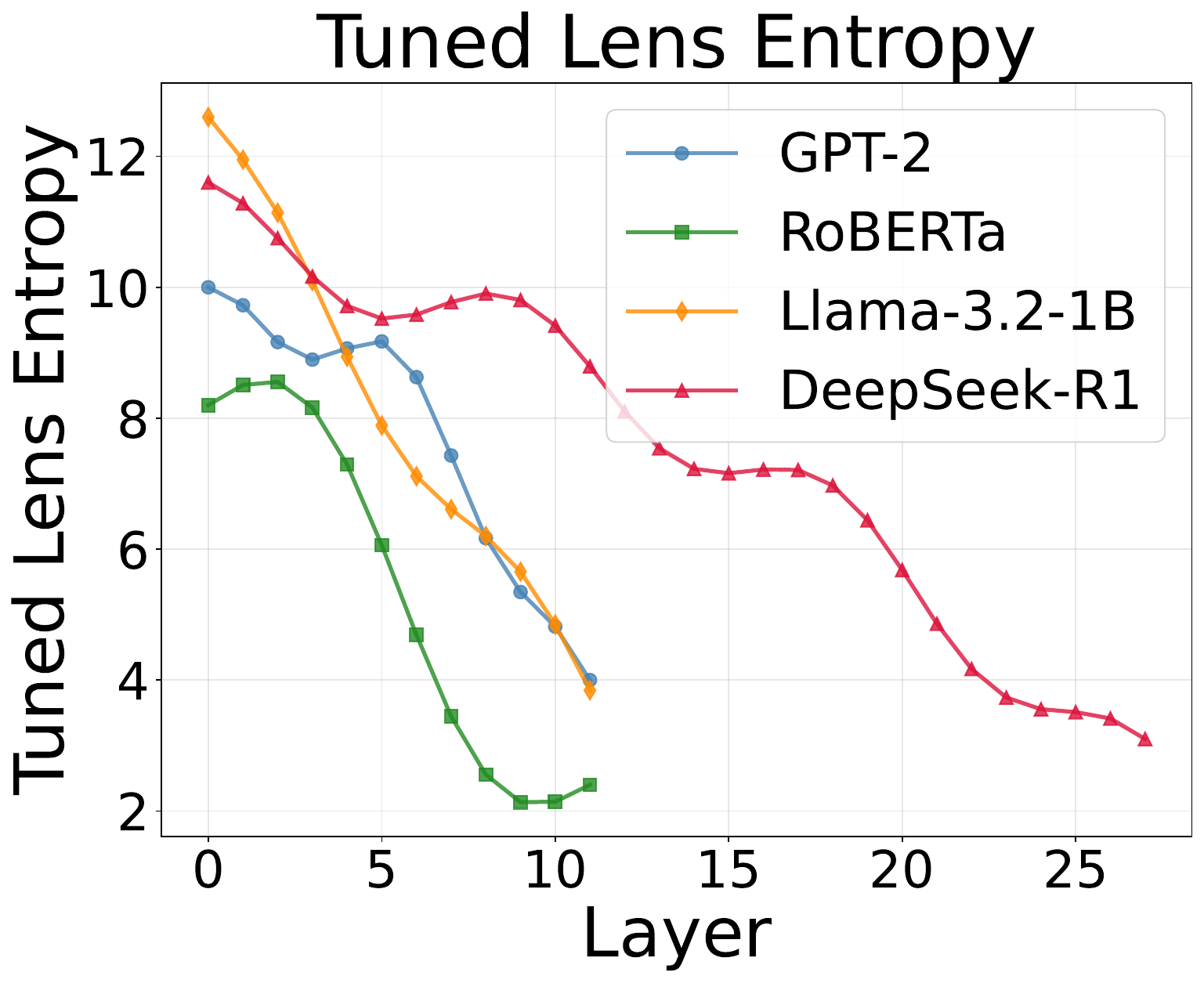}
\hfill
\includegraphics[width=0.19\textwidth]{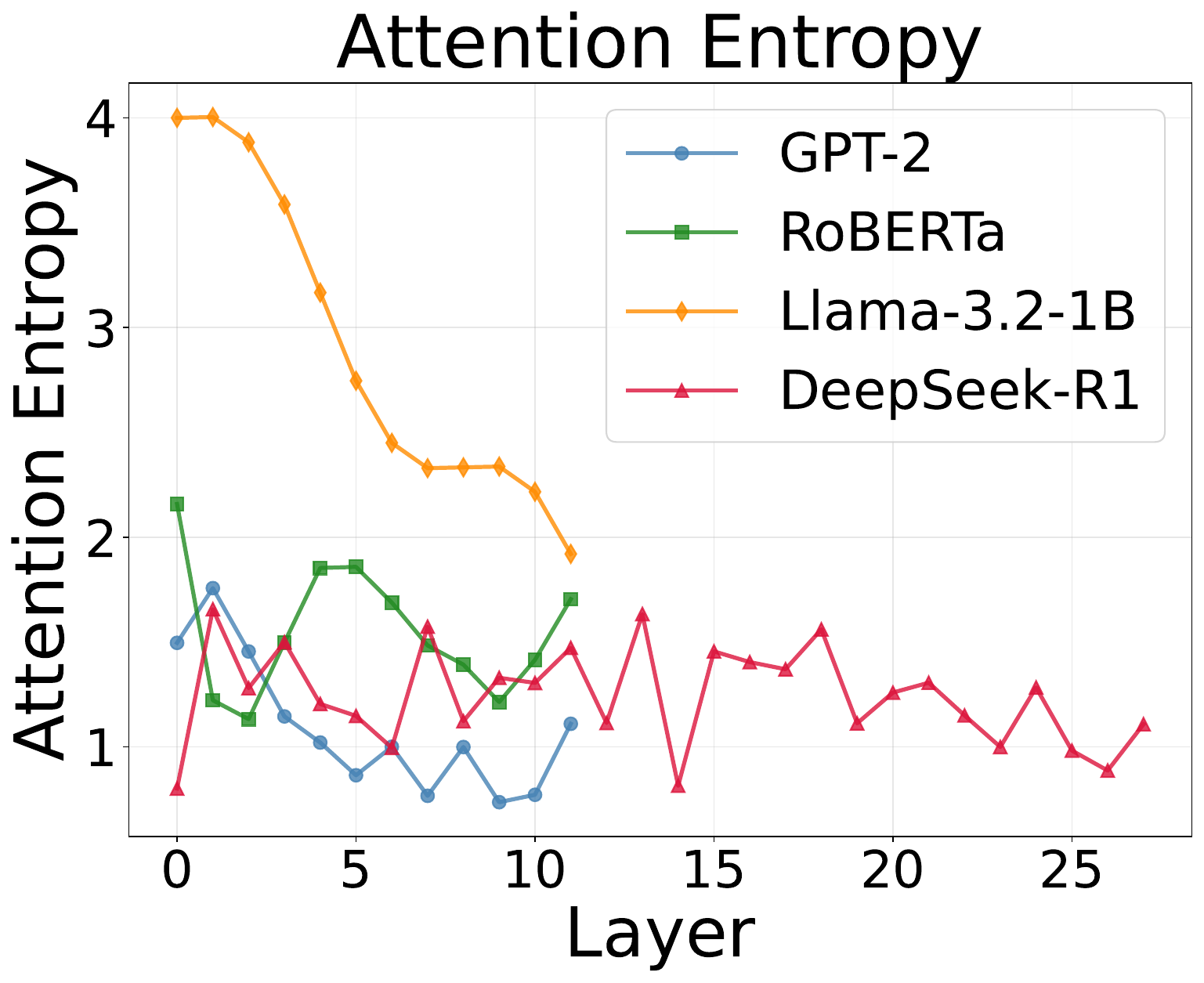}
\hfill
\includegraphics[width=0.19\textwidth]{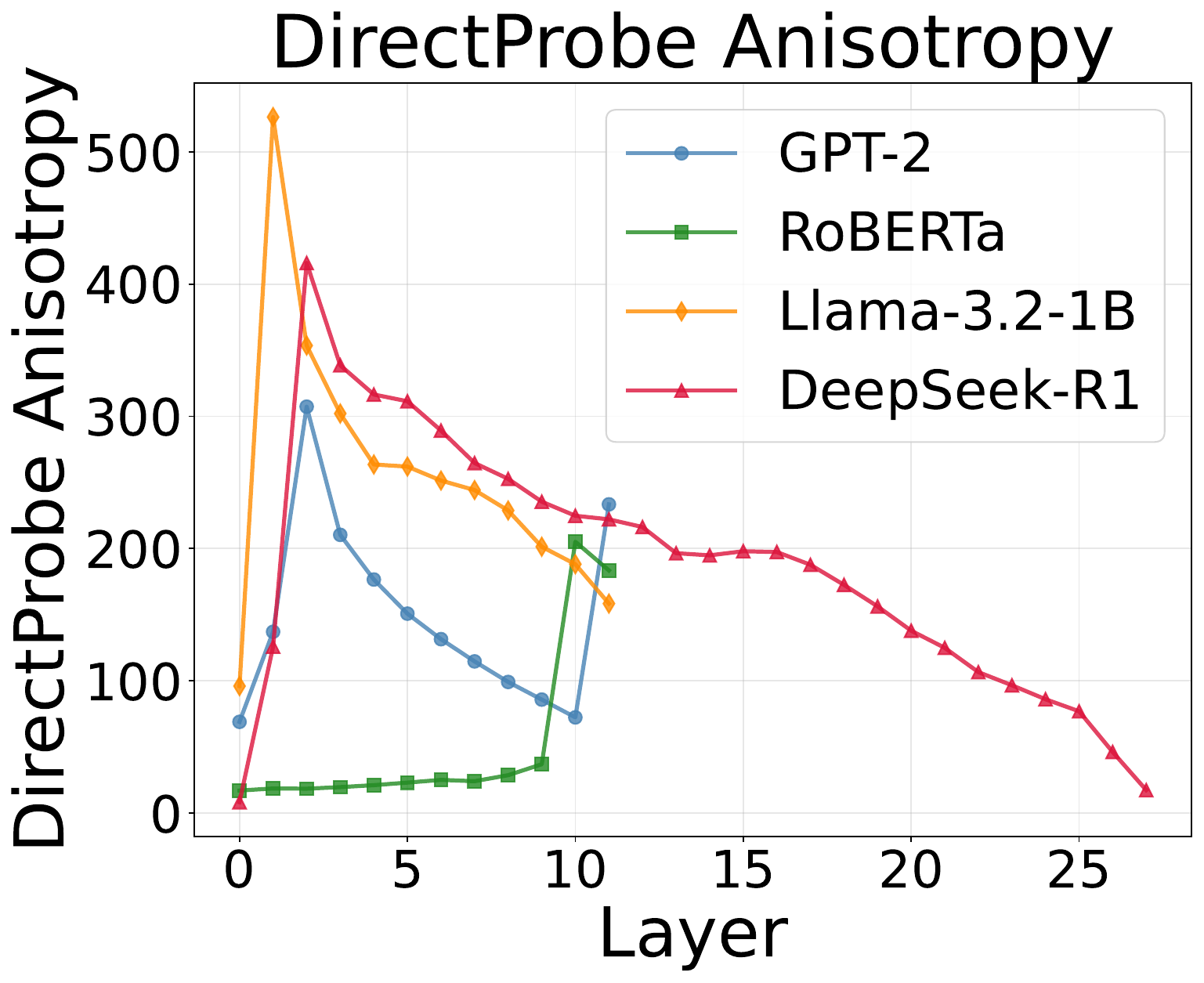}

\caption{Comparison of CAST with complementary methods across four architectures. From left to right: CAST Effective Rank reveals architecture-specific compression patterns, Logit Lens Entropy tracks prediction-space evolution, Tuned Lens Entropy captures refined prediction dynamics, Attention Entropy shows attention focusing patterns, and DirectProbe Anisotropy measures representation isotropy. The comparison demonstrates that CAST captures unique structural properties.}
\label{fig:baseline_comparison}
\end{figure*}

\begin{table}[t]
\centering
\caption{Layer-wise transformation metrics for GPT-2. Abbreviations: ER=Effective Rank, SDR=Spectral Decay Rate, TE=Transformation Entropy, AI=Anisotropy Index, IC=Information Concentration, RN=Residual Norm. Role assignments in the table are based on prior work~\citep{tenney2019bert,rogers2020primer}.}
\label{tab:layer_characterization}
\begin{tabular}{@{}c@{~}|@{~}c@{~}@{~}c@{~}@{~}c@{~}@{~}c@{~}@{~}c@{~}@{~}c@{~}|@{~}l@{}}
\hline
\textbf{Layer} & \textbf{ER} & \textbf{SDR} & \textbf{TE} & \textbf{AI} & \textbf{IC} & \textbf{RN} & \textbf{Role} \\
\hline
0 & 683 & 0.01 & 6.22 & 10.35 & -0.49 & 10.89 & Token$\rightarrow$Feature \\
1 & 740 & 0.00 & 6.47 & 21.02 & -0.28 & 8.92 & Feature Expansion \\
2 & 728 & 0.01 & 6.50 & 11.05 & -0.26 & 10.90 & Feature Expansion \\
3 & 323 & 0.03 & 5.77 & 4.56 & -0.61 & 15.33 & Syntax Analysis \\
4 & 357 & 0.03 & 5.87 & 3.77 & -0.57 & 15.85 & Syntax Analysis \\
5 & 384 & 0.03 & 5.94 & 3.42 & -0.54 & 16.83 & Semantic Core \\
6 & 419 & 0.03 & 6.03 & 3.35 & -0.50 & 18.52 & Semantic Core \\
7 & 468 & 0.03 & 6.13 & 3.13 & -0.45 & 20.19 & Context Integration \\
8 & 540 & 0.03 & 6.27 & 2.96 & -0.38 & 23.01 & Context Integration \\
9 & 628 & 0.02 & 6.42 & 2.74 & -0.29 & 25.36 & Context Integration \\
10 & 711 & 0.01 & 6.53 & 3.88 & -0.21 & 30.35 & Specialization \\
11 & 756 & 0.00 & 6.57 & 9.34 & -0.19 & 38.40 & Output Prep \\
\hline
\end{tabular}
\vspace{-2em}
\end{table}

\subsection{Complementary Analysis Methods}
\label{sec:complementary_methods}

We compare CAST with complementary methods that illuminate distinct facets of transformer processing:

\textbf{Logit Lens}~\citep{nostalgebraist2020logit} projects intermediate layer representations to vocabulary space through the language model head, revealing how predictions evolve across depth. Early layers produce noisy predictions that progressively refine into confident outputs in deeper layers.

\textbf{Tuned Lens}~\citep{belrose2023eliciting} improves logit lens by learning affine transformations that align intermediate representations with the final layer before projection. This reduces architectural misalignment artifacts and provides clearer insights into iterative prediction refinement.

\textbf{DirectProbe}~\citep{zhou2021directprobe,razzhigaev2024shape} analyzes representation geometry without auxiliary classifiers, using SVD to measure anisotropy and dimensionality. The method reveals representations become increasingly anisotropic with depth, concentrating in task-specific subspaces.

\begin{figure*}[t]
\centering
\begin{subfigure}[b]{0.245\textwidth}
    \centering
    \includegraphics[width=\textwidth]{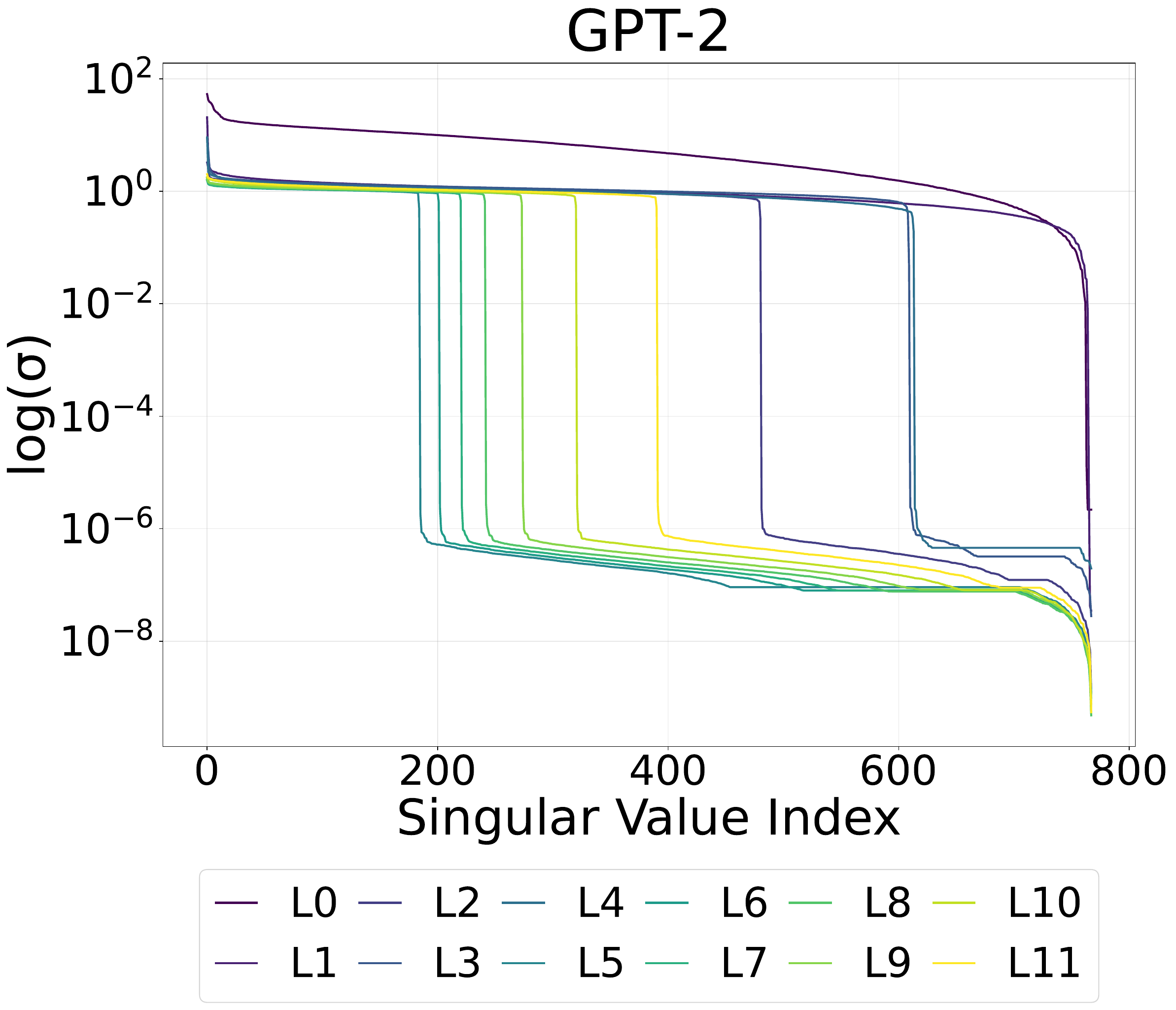}
\end{subfigure}
\hfill
\begin{subfigure}[b]{0.245\textwidth}
    \centering
    \includegraphics[width=\textwidth]{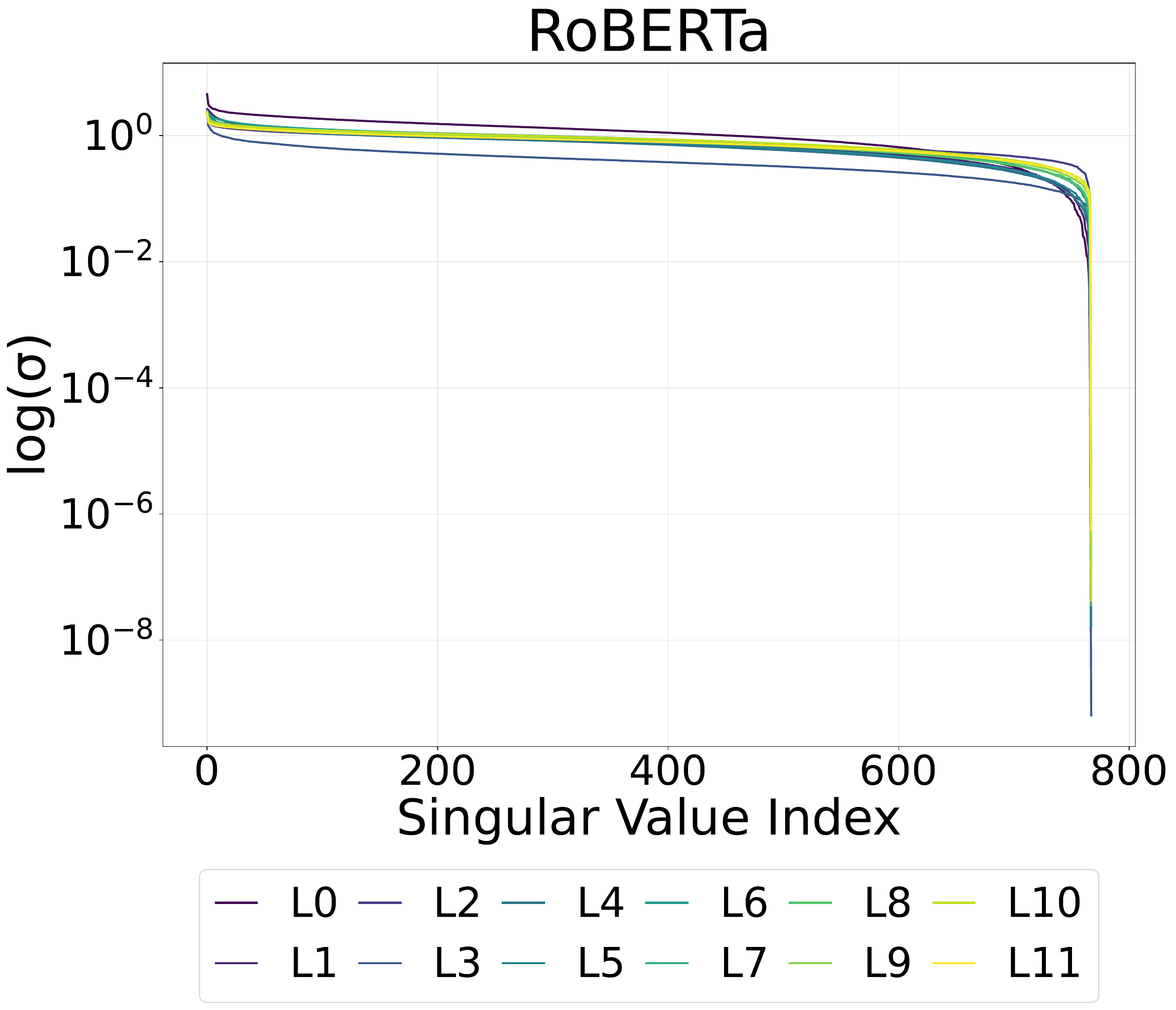}
\end{subfigure}
\hfill
\begin{subfigure}[b]{0.245\textwidth}
    \centering
    \includegraphics[width=\textwidth]{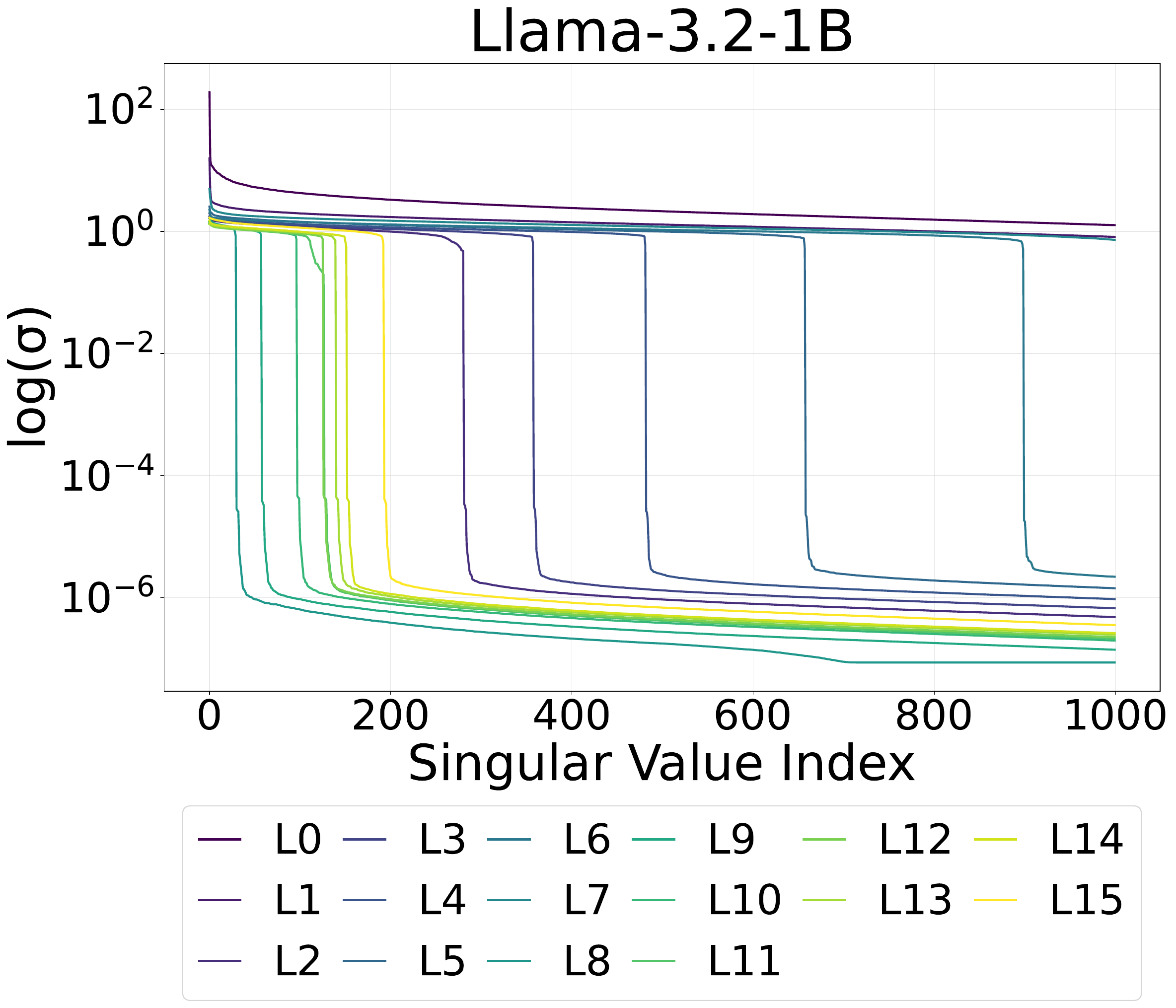}
\end{subfigure}
\hfill
\begin{subfigure}[b]{0.245\textwidth}
    \centering
    \includegraphics[width=\textwidth]{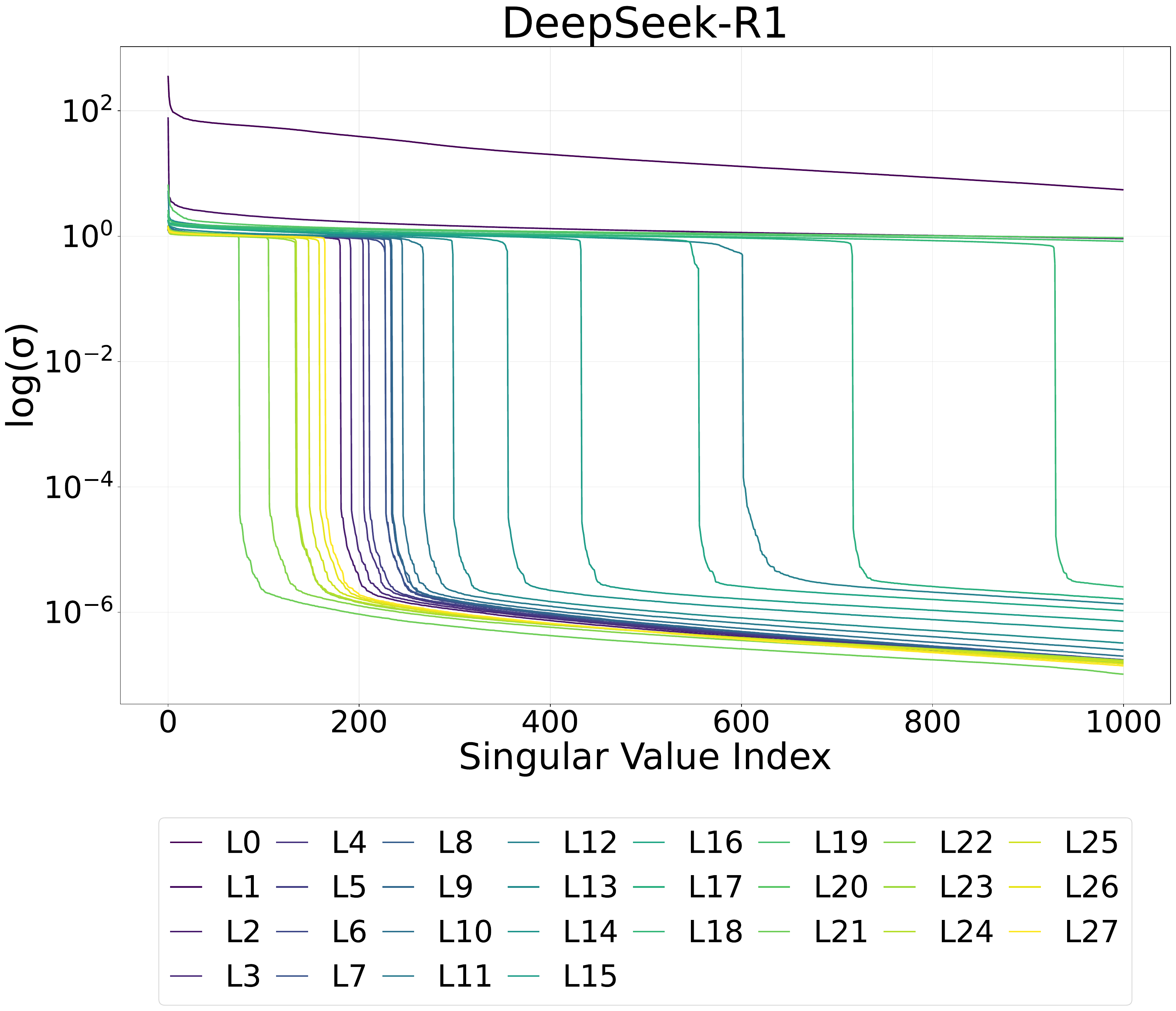}
\end{subfigure}
\caption{Singular value distributions across layers. GPT-2 shows pronounced compression at middle layers with steep spectral decay, RoBERTa maintains gentle decay curves preserving information, Llama exhibits moderate compression patterns, while DeepSeek-R1 shows sustained high-rank representations across its layers with gradual spectral evolution.}
\label{fig:svd_dist}
\end{figure*}

\begin{figure}[t]
\centering
\includegraphics[width=0.95\textwidth]{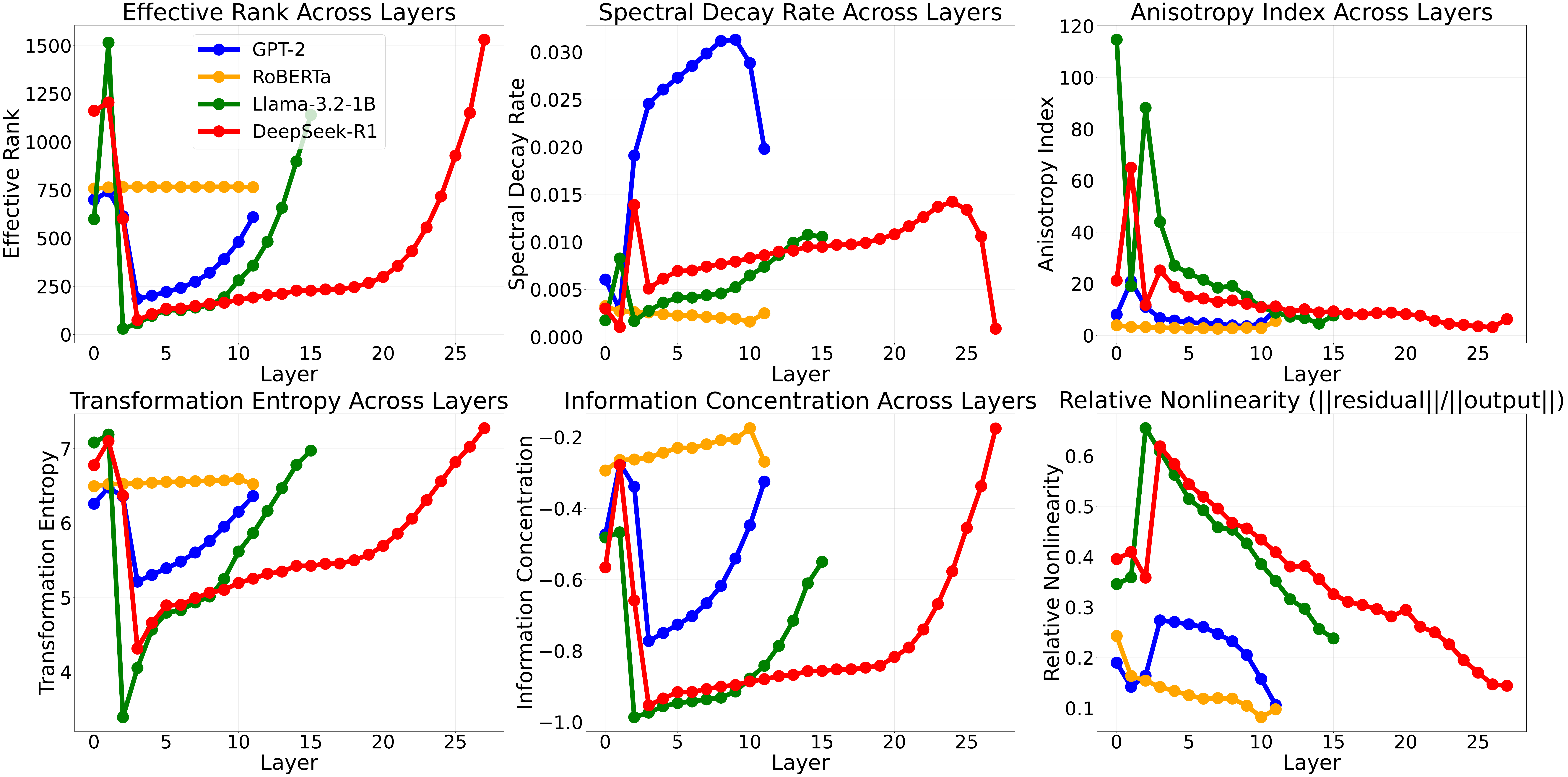}
\caption{Comparison of CAST metrics. Decoder models exhibit compression-expansion cycles while encoder models maintain consistent high-rank processing.}
\label{fig:cross_arch}
\end{figure}

\textbf{Attention Entropy}~\citep{vig2019analyzing} quantifies attention concentration by computing entropy across attention weights at each layer. Low entropy indicates focused attention on specific tokens, while high entropy suggests uniform attention distribution.

\subsection{Layer Characterization Analysis}
\label{sec:layer_characterization}

To understand how transformer layers specialize in computational roles, we analyze GPT-2 layer-wise transformations in \Cref{tab:layer_characterization}. We extract hidden states from layer transitions, estimate transformation matrices using Moore-Penrose pseudoinverse, compute six spectral metrics per layer. Role assignments follow prior work~\citep{tenney2019bert,rogers2020primer}. Results: (1) effective rank follows U-shaped trajectory characteristic of autoregressive decoders—expanding early, compressing middle, re-expanding late—demonstrating information bottleneck where models extract features, compress for abstraction, expand for task-specific computation, consistent with information-theoretic analyses~\citep{tishby2015deep,schwartz2017opening} and transformer findings~\citep{voita2019bottom,tenney2019bert}; (2) spectral decay rate triples during compression, suggesting dimensionality reduction facilitates abstract linguistic pattern extraction from high-dimensional spaces, aligning with representation learning~\citep{bengio2013representation} and specialization studies~\citep{rogers2020primer,kovaleva2019revealing}; (3) residual norm increases monotonically, revealing deeper layers require increasingly nonlinear transformations for complex semantic relationships beyond linear mappings; (4) three functional phases emerge—feature expansion, semantic compression, output specialization—with distinct spectral signatures, confirming transformers implement hierarchical processing similar to classical NLP systems, corroborating probing~\citep{tenney2019bert,hewitt2019structural} and mechanistic interpretability~\citep{elhage2021mathematical,olah2020zoom}.

\subsection{Method Comparison with Complementary Methods}

To position CAST within transformer interpretability methods, we compare with complementary approaches from \Cref{sec:complementary_methods} across four architectures in \Cref{fig:baseline_comparison}. Observations: (1) CAST Effective Rank uniquely captures architecture-specific transformation dynamics—GPT-2 shows dramatic compression at middle layers then recovery, RoBERTa maintains consistently high rank reflecting bidirectional processing, Llama shows gradual compression, DeepSeek-R1 demonstrates sustained high-rank processing with mild compression, making CAST the only method distinguishing autoregressive compression from bidirectional preservation behaviors; (2) Logit Lens and Tuned Lens Entropy show monotonic decrease across architectures, demonstrating layers progressively reduce entropy transforming uncertain representations into confident predictions; (3) Attention Entropy displays high variability—fluctuating patterns in GPT-2 and Llama, structured evolution in RoBERTa, irregular oscillations in DeepSeek-R1—suggesting attention mechanisms are influenced by training dynamics not architectural principles; (4) DirectProbe Anisotropy reveals dramatic scale differences—early-layer peaks in GPT-2, late-layer increases in RoBERTa, high early-layer values in DeepSeek-R1—showing it focuses on representation geometry not computational dynamics; (5) projection-based methods focus on output space evolution, geometric methods examine static properties, while CAST measures transformation complexity revealing how architectures implement distinct information processing strategies. CAST and complementary methods provide different perspectives—CAST offers insights into transformation dynamics complementing existing approaches.

\subsection{Singular Value Distribution Analysis}

To understand information processing across transformer architectures, we conduct singular value distribution analysis comparing GPT-2, RoBERTa, Llama, and DeepSeek-R1 shown in \Cref{fig:svd_dist}. We compute SVD for each layer's transformation matrix and plot distributions on log scale, revealing architecture-specific spectral decay patterns. Visualization shows how layers compress or preserve information through singular value spectra. For decoder models, we observe sharp singular value decrease at transition points: decay begins gradually, steepens in middle layers, then recovers. Results show (1) all three decoder models (GPT-2, Llama, DeepSeek-R1) exhibit compression-expansion patterns with dramatic spectral changes—early layers maintain broad spectra that collapse at middle layers before recovering, demonstrating compression bottleneck is fundamental property of autoregressive architectures optimized for next-token prediction, appearing across scales and training paradigms—validating compression-expansion phenomenon in transformer analysis~\citep{tenney2019bert,voita2019bottom}; (2) RoBERTa displays consistent singular value distributions across layers with gentle decay curves maintaining magnitude at high indices, revealing bidirectional encoders preserve information throughout depth to support downstream tasks without committing to specific predictions; (3) visualization confirms spectral properties reflect architectural objectives rather than model-specific artifacts, validating transformation matrices capture fundamental differences between autoregressive and bidirectional information processing strategies.

\subsection{CAST Metrics Across Architectures}

We apply CAST analysis to four architectures—GPT-2, RoBERTa, Llama, and DeepSeek-R1—measuring six transformation metrics across layers shown in \Cref{fig:cross_arch}. Results reveal how layers specialize in distinct computational roles: (1) Effective Rank shows decoder models compress information dimensions in middle layers (low rank for abstraction)~\citep{ansuini2019intrinsic,razzhigaev2024shape} then expand in later layers (high rank for output specialization), while encoders maintain high dimensionality throughout preserving bidirectional context; (2) Spectral Decay Rate increases sharply during compression phases where layers aggressively reduce singular values, indicating dimensional reduction for feature extraction; (3) Anisotropy Index reveals early layers process inputs uniformly (low anisotropy) while middle layers develop strong directional preferences (high anisotropy) for linguistic patterns; (4) Transformation Entropy decreases in compression phases as layers concentrate processing power into fewer dominant directions, then increases as layers become complex for output generation; (5) Information Concentration peaks at bottleneck layers where transformation power is concentrated in few singular values, showing aggressive feature selection; and (6) Relative Nonlinearity increases with depth as layers require complex transformations to handle abstract semantic relationships beyond linear mappings.

\begin{figure}[t]
\centering
\includegraphics[width=0.9\textwidth]{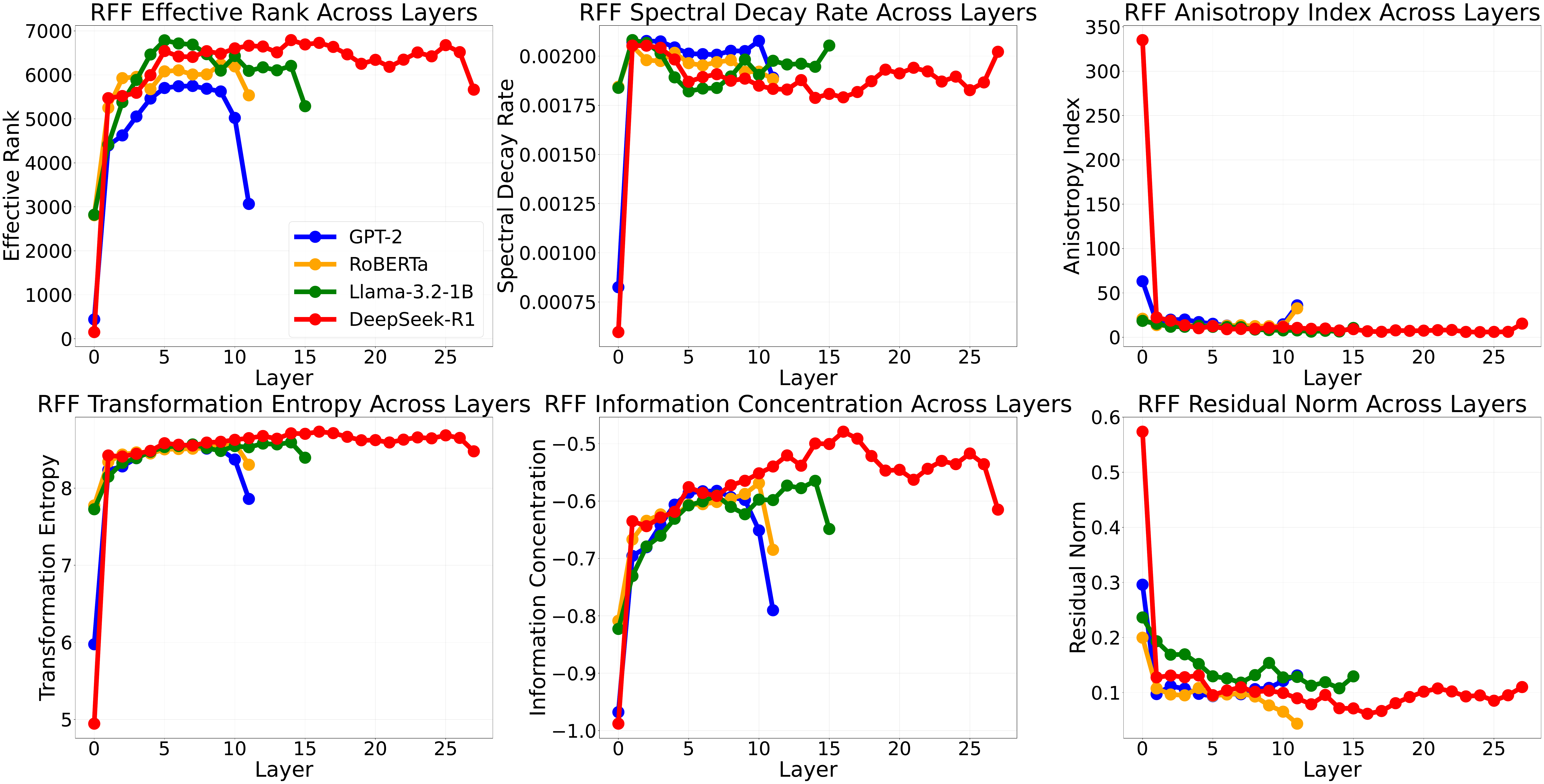}
\caption{Random Fourier Features analysis reveals complementary patterns to linear analysis with consistently lower residual norms, demonstrating enhanced nonlinear transformation modeling capabilities.}
\label{fig:rff_metrics}
\end{figure}

\begin{figure}[t]
\centering
\begin{subfigure}[b]{0.43\textwidth}
    \centering
    \includegraphics[height=0.9\textwidth]{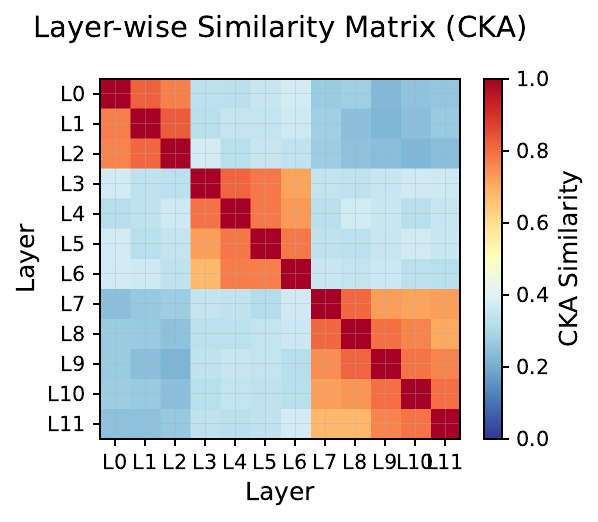}
    \caption{Layer-wise similarity (CKA)}
\end{subfigure}
\hfill
\begin{subfigure}[b]{0.43\textwidth}
    \centering
    \includegraphics[height=0.9\textwidth]{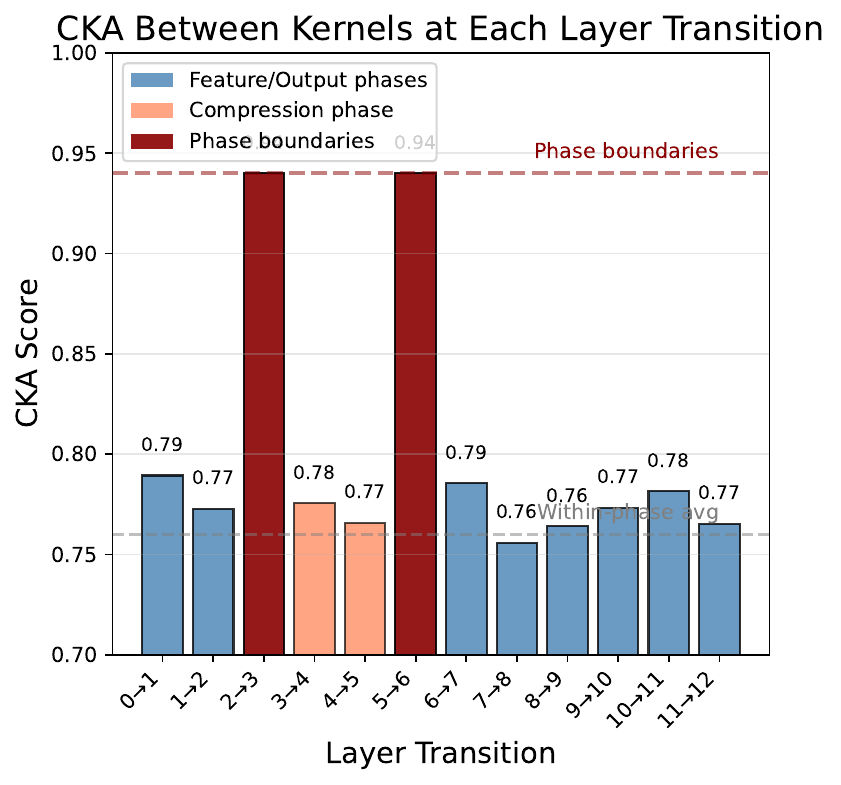}
    \caption{CKA at layer transitions}
\end{subfigure}
\caption{Kernel analysis revealing complementary transformation perspectives. (a) CKA similarity matrix showing three distinct functional phases. (b) Layer transition CKA peaks at phase boundaries.}
\label{fig:kernel_analysis}
\vspace{-1em}
\end{figure}

\subsection{Random Fourier Features Analysis}

We apply Random Fourier Features (RFF) to analyze nonlinear transformations that complement our linear analysis. \Cref{fig:rff_metrics} shows six metrics computed in RFF space using RBF kernels. Comparing RFF with linear analysis reveals complementary insights: (1) Effective ranks show opposite trends—while linear analysis revealed compression-expansion cycles with low-rank middle layers, RFF maintains consistently high effective ranks throughout, suggesting RFF captures nonlinear spectral characteristics invisible to direct linear methods; (2) Architectural differences persist in kernel space—decoder models maintain distinctive patterns while encoders show stable high ranks, confirming fundamental differences exist but with enhanced dimensional richness; (3) Spectral decay rates are consistently higher than linear values, with distinctive first-layer drops revealing that nonlinear transformations concentrate power in fewer principal directions despite higher effective ranks; (4) Anisotropy patterns show extreme first-layer values before stabilizing to moderate levels, contrasting with linear analysis where anisotropy gradually increases with depth—this suggests kernel transformations exhibit strong initial directional bias that gets regularized in deeper layers; (5) Information concentration patterns are inverted compared to linear analysis, demonstrating complementary behavior where kernel transformations distribute information more uniformly across principal components; (6) RFF achieves consistently lower residual norms compared to linear analysis, demonstrating that kernel methods better model actual transformation complexity. These findings demonstrate that RFF specifically captures nonlinear transformation properties that linear analysis cannot reveal, showing transformer layers implement rich nonlinear transformations while preserving dimensional complexity.

\subsection{Layer Similarity Analysis}

To explore whether linear analysis captures transformation structure, we conduct kernel analysis shown in \Cref{fig:kernel_analysis}. Using RBF kernels, we analyze layer similarity patterns through Centered Kernel Alignment (CKA). Analysis demonstrates key insights: (1) CKA similarity matrices partition layers into three functional phases shown in panels (a) and (b)—early feature extraction layers exhibit high intra-phase similarity, middle compression layers form coherent block with distinct characteristics, and later specialization layers show unified behavior, validating our identified three-phase architecture; (2) layer transitions between phases show clear boundaries in CKA values, with the most pronounced changes occurring at phase transitions, confirming that these phases represent fundamentally different computational operations; and (3) the block-diagonal structure of the CKA matrix reveals that transformer layers implement a systematic progression of information processing, with each phase maintaining internal consistency while being distinct from other phases.

\section{Conclusion}

We propose CAST, a novel analytical framework that complements existing interpretability methods by providing transformation-centric insights into transformer layer functions through direct matrix estimation and spectral analysis. Our framework uniquely captures the realized computational operations during forward passes, revealing architectural patterns invisible to probe-based and projection methods. Experimental analysis across GPT-2, RoBERTa, Llama, and DeepSeek-R1 reveals fundamental differences in information processing strategies: decoder-only models exhibit compression-expansion cycles optimized for sequential prediction, while encoder-only models maintain high-rank processing throughout their depth for bidirectional understanding. Multi-kernel analysis further demonstrates that middle compression layers involve the strongest nonlinear transformations, with consistent patterns observed across different architectures and sample sizes. These findings provide practical guidance for layer pruning, architecture design, and training optimization. CAST provides mathematical tools for understanding transformer computations and opens new directions for interpretable language model development.

%% file: appendix.tex
\clearpage
\onecolumn
\setcounter{section}{1}
\makeatletter
\let\oldsection\section
\renewcommand{\section}[1]{%
  \refstepcounter{section}%
  \oldsection*{\Alph{section}. #1}%
  \addcontentsline{toc}{section}{\Alph{section}. #1}%
}
\makeatother
\begin{center}
  {\LARGE \textbf{Appendix. Supplementary Material}}
\end{center}

\section{Limitations}

While CAST provides valuable insights into transformer layer functions, several limitations should be acknowledged. Importantly, CAST is designed as a complementary approach to existing interpretability methods rather than a replacement or superior alternative—it offers a transformation-centric perspective that works alongside probe-based methods, attention visualization, and mechanistic interpretability to provide a more complete understanding of transformer behavior. First, our linear approximation approach, though effective for capturing primary transformation patterns, may not fully capture the complete nonlinear dynamics within transformer layers, particularly the complex interactions between attention mechanisms and feed-forward networks. Second, our analysis focuses on representation-level transformations and does not directly examine the internal computations within individual layer components such as multi-head attention or position-wise feed-forward networks, areas where existing mechanistic interpretability methods excel. Third, the framework's reliance on sufficient sample sizes for stable pseudoinverse computation may limit its applicability to scenarios with limited data availability. Fourth, while we validate our approach across four diverse architectures, the generalizability to emerging transformer variants and next-generation architectures requires further investigation. Finally, our spectral analysis provides insights into transformation structure but does not directly address the semantic interpretability of the identified patterns, which benefits from integration with existing probing and visualization techniques for complete understanding.

\section{Impact Statement}

This work aims to advance the interpretability of transformer-based language models through mathematical analysis of layer-wise transformations. The CAST framework has several potential positive impacts: it provides researchers and practitioners with new tools for understanding model behavior, enables more informed decisions about model architecture design and optimization, and contributes to the broader goal of making AI systems more transparent and interpretable. The insights from our analysis can guide practical applications such as model compression, efficient training strategies, and architecture design principles.

However, we acknowledge that increased interpretability tools could potentially be misused. While our methods are designed for defensive analysis and understanding, any interpretability technique could theoretically be leveraged for adversarial purposes, such as identifying model vulnerabilities or developing more sophisticated attacks. Additionally, the computational insights provided by CAST could inform the development of more efficient models, which might accelerate AI capabilities in ways that require careful consideration of broader societal impacts.

We emphasize that CAST is intended as a research tool for improving our understanding of transformer architectures and should be used responsibly within appropriate ethical frameworks. The development of interpretable AI systems is crucial for ensuring their safe and beneficial deployment across various applications.

\section{Additional Experimental Details}

\subsection{Statistical Validation with Bootstrap Confidence Intervals}

\begin{figure*}[ht]
\centering
\includegraphics[width=\textwidth]{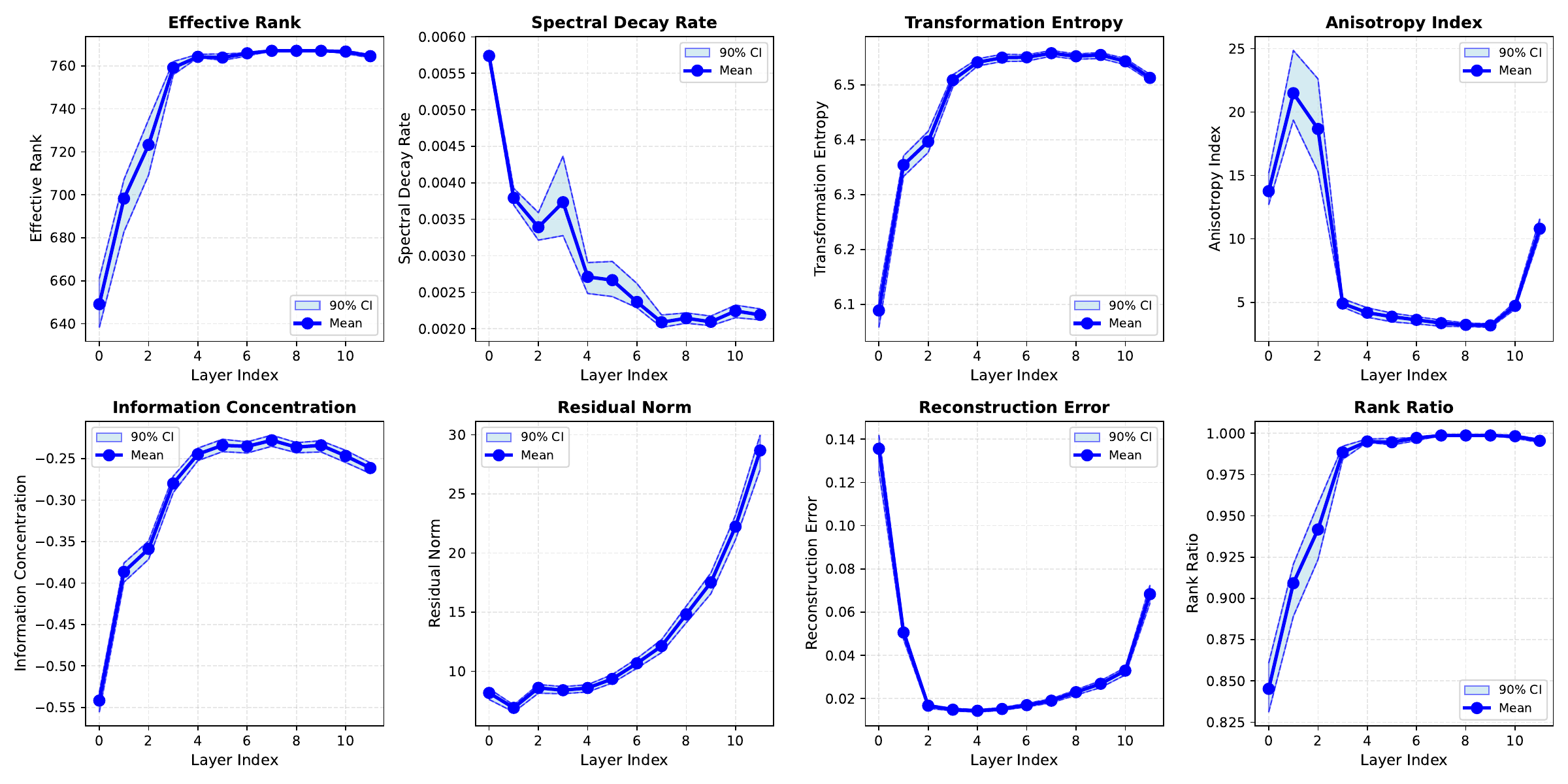}
\caption{Bootstrap confidence intervals for eight key CAST metrics across layers. Each subplot shows the mean (blue line) and 95\% confidence interval (shaded region) computed from 20 bootstrap iterations. The selected metrics include effective rank, spectral decay rate, transformation entropy, anisotropy index, information concentration, residual norm, reconstruction error, and rank ratio. The analysis demonstrates high metric stability across resampling, with narrow confidence intervals confirming the statistical reliability of our measurements.}
\label{fig:bootstrap_ci}
\end{figure*}

To assess the statistical reliability of our CAST metrics, we conduct bootstrap analysis with $20$ iterations as shown in \Cref{fig:bootstrap_ci}. Using a fixed dataset of $2000$ sequences, we perform bootstrap sampling with replacement to compute $95\%$ confidence intervals for eight key CAST metrics across layers. This analysis quantifies metric stability and measurement uncertainty independent of sample size effects. We can observe from the results that (1) core metrics show high stability with narrow confidence intervals—effective rank has $95\%$ CI width of $\pm 5.9$ ($21.7\%$ of mean), transformation entropy $\pm 0.24$ ($7.5\%$ of mean), and rank ratio $\pm 0.01$ ($25\%$ of mean), which demonstrates that our spectral analysis captures stable properties of the transformation matrices rather than noise or sampling artifacts; (2) condition number exhibits the highest variability with CI spanning two orders of magnitude, which reflects its extreme sensitivity to small singular values near machine precision, making it a poor metric for layer characterization despite its theoretical importance; (3) nonlinearity metrics (residual norm, reconstruction error) show extremely tight bounds ($\pm 0.001$), which confirms that the nonlinear components of transformer layers are highly consistent across different data samples, supporting their use as reliable indicators of layer function; (4) layer-specific patterns are preserved across all bootstrap iterations—the compression bottleneck at layers $3$-$6$ appears in every sample with consistent magnitude, which proves that the architecture-specific patterns (compression-expansion cycles in decoders, consistent high-rank processing in encoders) are fundamental properties of their respective architectures rather than statistical fluctuations; and (5) the bootstrap analysis validates that our main findings are statistically robust, which provides the necessary confidence to draw conclusions about transformer information processing from finite data samples.

\subsection{Sample Size Sensitivity Analysis}

\begin{figure*}[t]
\centering
\begin{subfigure}[b]{0.45\textwidth}
    \centering
    \includegraphics[width=0.8\textwidth]{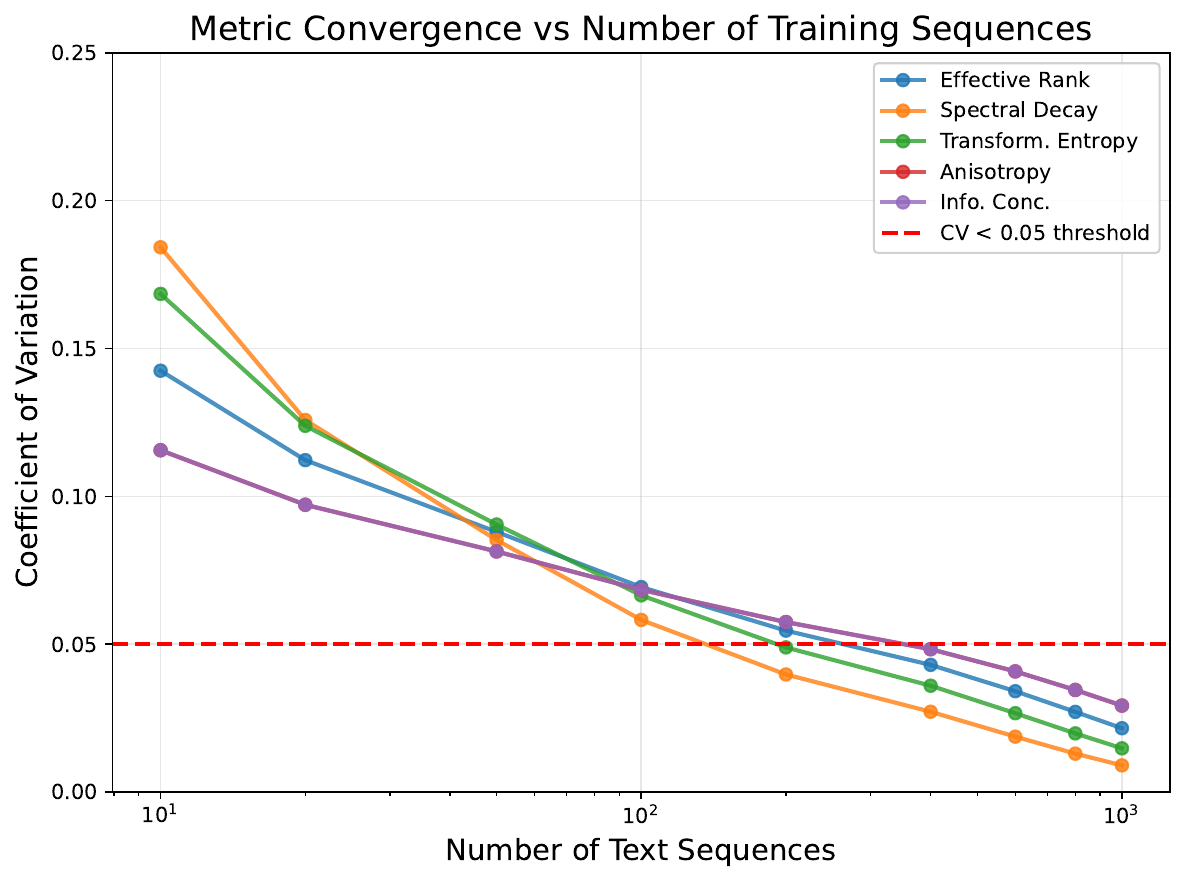}
    \caption{Metric convergence}
\end{subfigure}
\hfill
\begin{subfigure}[b]{0.45\textwidth}
    \centering
    \includegraphics[width=0.8\textwidth]{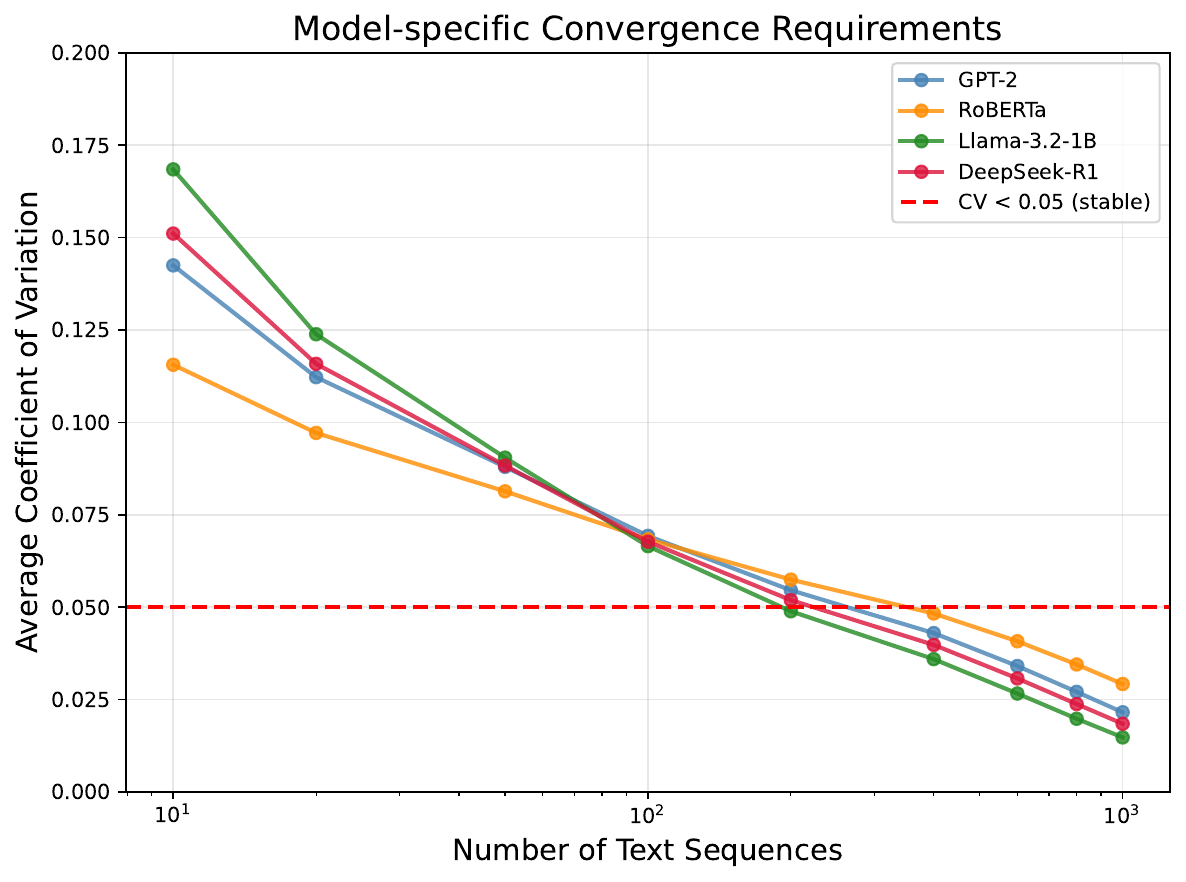}
    \caption{Model-specific convergence}
\end{subfigure}
\caption{Sample size sensitivity analysis for CAST metrics. (a) Convergence of different CAST metrics showing that effective rank converges fastest while spectral decay and transformation entropy require more sequences. (b) Model-specific convergence requirements demonstrating that Llama-3.2 requires more samples for stability compared to RoBERTa, with GPT-2 and DeepSeek-R1 showing intermediate behavior.}
\label{fig:sample_size_analysis}
\end{figure*}

To establish minimum data requirements for reliable CAST analysis, we conduct sample size sensitivity experiments as shown in \Cref{fig:sample_size_analysis}. We analyze metric stability across different numbers of text sequences, computing coefficient of variation (CV = standard deviation / mean) with multiple random seeds per configuration. We can observe from the results that (1) all experimental metrics demonstrate consistent sample size-dependent stability, with larger sample sizes leading to more stable estimates across all measured quantities; (2) for all metrics, when sample size exceeds a critical threshold, the coefficient of variation drops substantially and plateaus at low levels, demonstrating clear convergence behavior that validates the statistical reliability of our measurements; and (3) across all tested models, convergence patterns emerge consistently when sample size reaches sufficient levels, though different architectures exhibit varying sensitivity requirements, with some models achieving stability with smaller datasets while others require more extensive sampling to reach equivalent measurement precision.

\subsection{Matrix Estimation Method Comparison}

To validate our choice of Moore-Penrose pseudoinverse for transformation estimation, we conduct systematic comparison with ridge regression~\citep{hoerl1970ridge}, elastic net~\citep{zou2005regularization}, and truncated SVD~\citep{golub2013matrix} as shown in \Cref{tab:matrix_methods}. We can observe from the results that (1) pseudoinverse achieves minimal reconstruction error while preserving true effective rank patterns, whereas regularized methods inflate rank estimates and mask compression patterns; (2) the high condition number in pseudoinverse reflects the complex, high-dimensional nature of transformer data, and the unregularized approach better captures original layer behaviors compared to methods that artificially smooth singular value distributions; and (3) only pseudoinverse faithfully preserves the spectral properties necessary for identifying layer specialization phases.

\begin{table}[t]
\centering
\caption{Matrix estimation method comparison}
\label{tab:matrix_methods}
\begin{tabular}{@{}l|@{~}c@{~}@{~}c@{~}@{~}c@{~}@{~}c@{~}@{~}c@{}}
\hline
 & \textbf{Recon.} &  & \textbf{Eff.} &  & \textbf{Time} \\
\textbf{Method} & \textbf{Error} & \textbf{Condition} & \textbf{Rank} & \textbf{Decay} & \textbf{(s)} \\
\hline
Moore-Penrose & 0.0007 & $3.72 \times 10^8$ & 43.9 & 0.007 & 1.59 \\
Ridge (L2) & 0.0011 & $4.08 \times 10^7$ & 86.8 & 0.007 & 1.71 \\
Elastic Net & 0.0011 & $1.46 \times 10^7$ & 52.5 & 0.006 & 1.63 \\
Truncated SVD & 0.4365 & $3.23 \times 10^8$ & 39.6 & 0.007 & 2.77 \\
\hline
\end{tabular}
\end{table} 

\subsection{Implementation Details}

All experiments use PyTorch 2.0 with batch size 32, sequence length 512 tokens, random seed 42, NVIDIA A6000 40GB GPUs, and FP32 precision for transformation estimation.

\subsection{Dataset Preprocessing}

WikiText-103 sequences are tokenized using model-specific tokenizers, truncated to 512 tokens, filtered to remove sequences under 100 tokens, and randomly sampled with stratification by length to ensure diverse representation.

\subsection{RFF Spectral Distribution Analysis}

\begin{figure*}[ht]
\centering
\begin{subfigure}[b]{0.245\textwidth}
    \centering
    \includegraphics[width=\textwidth]{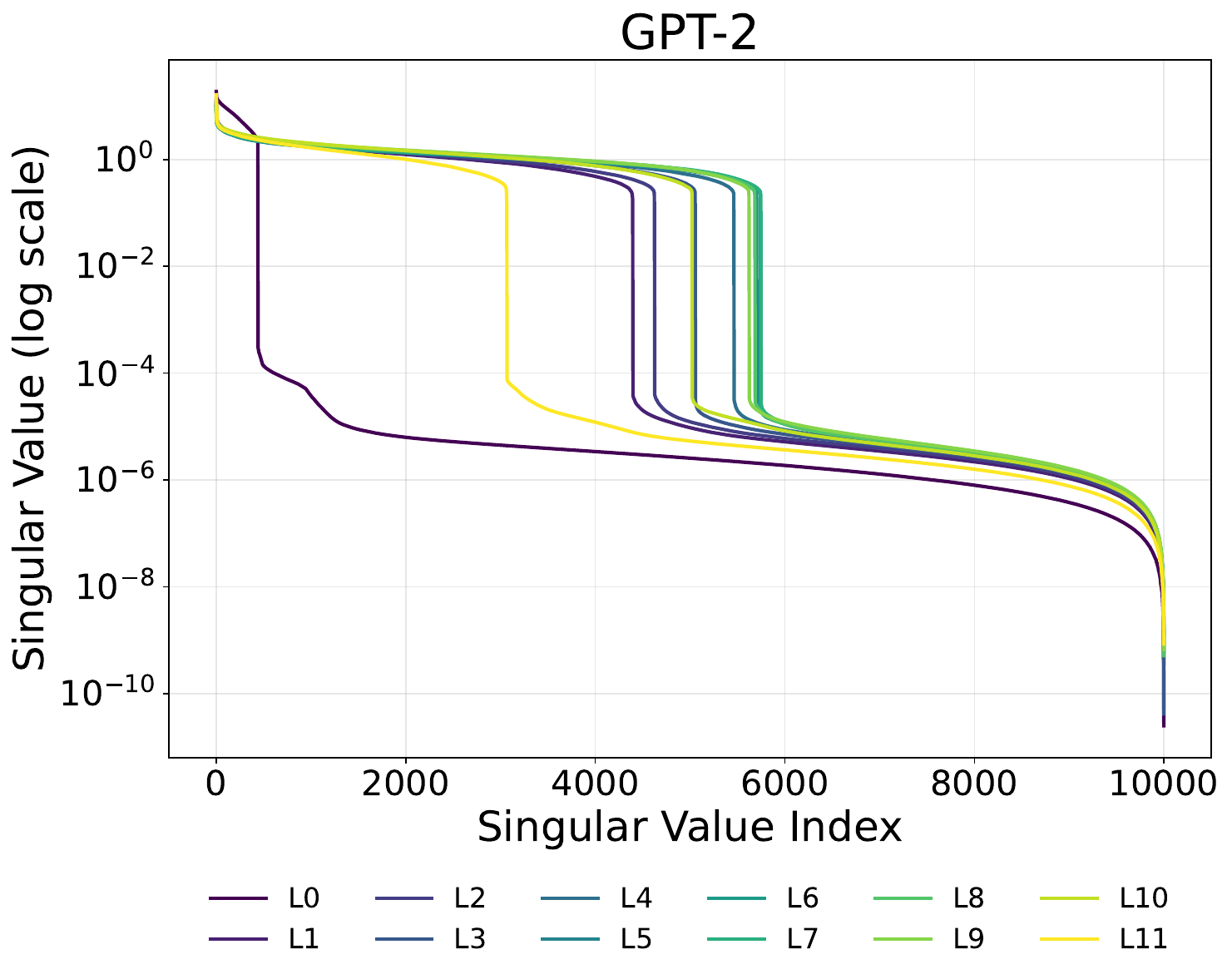}
\end{subfigure}
\hfill
\begin{subfigure}[b]{0.245\textwidth}
    \centering
    \includegraphics[width=\textwidth]{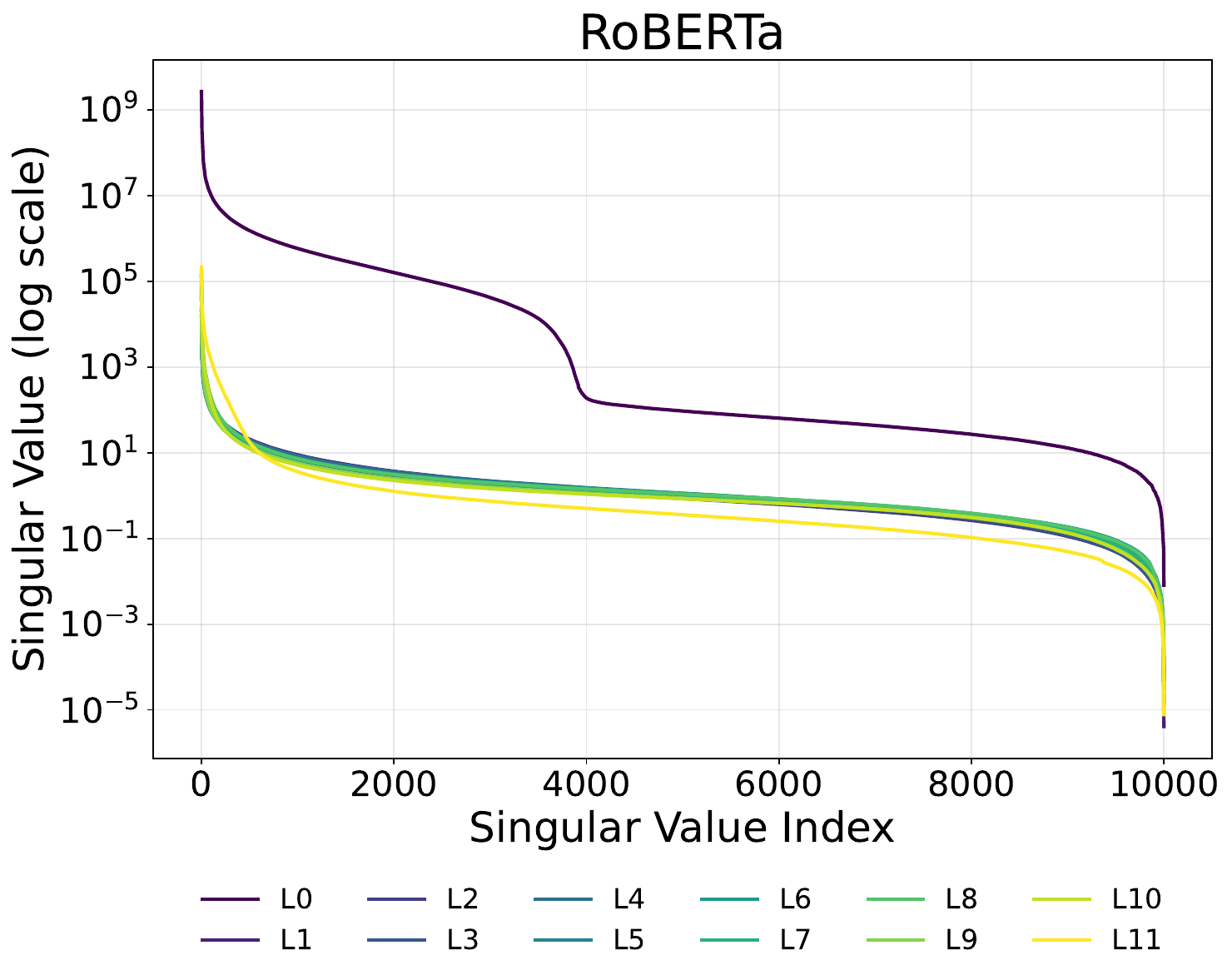}
\end{subfigure}
\hfill
\begin{subfigure}[b]{0.245\textwidth}
    \centering
    \includegraphics[width=\textwidth]{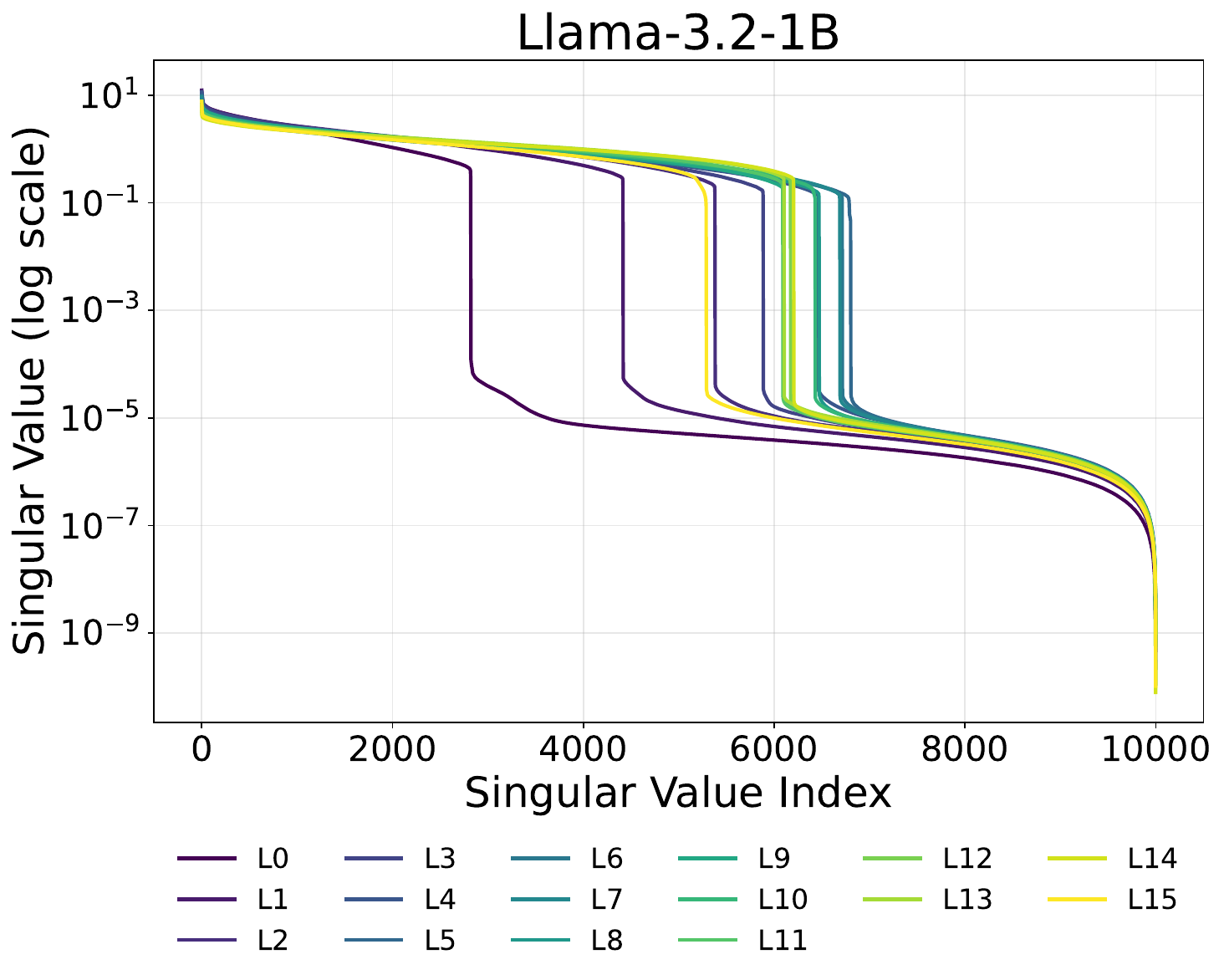}
\end{subfigure}
\hfill
\begin{subfigure}[b]{0.245\textwidth}
    \centering
    \includegraphics[width=\textwidth]{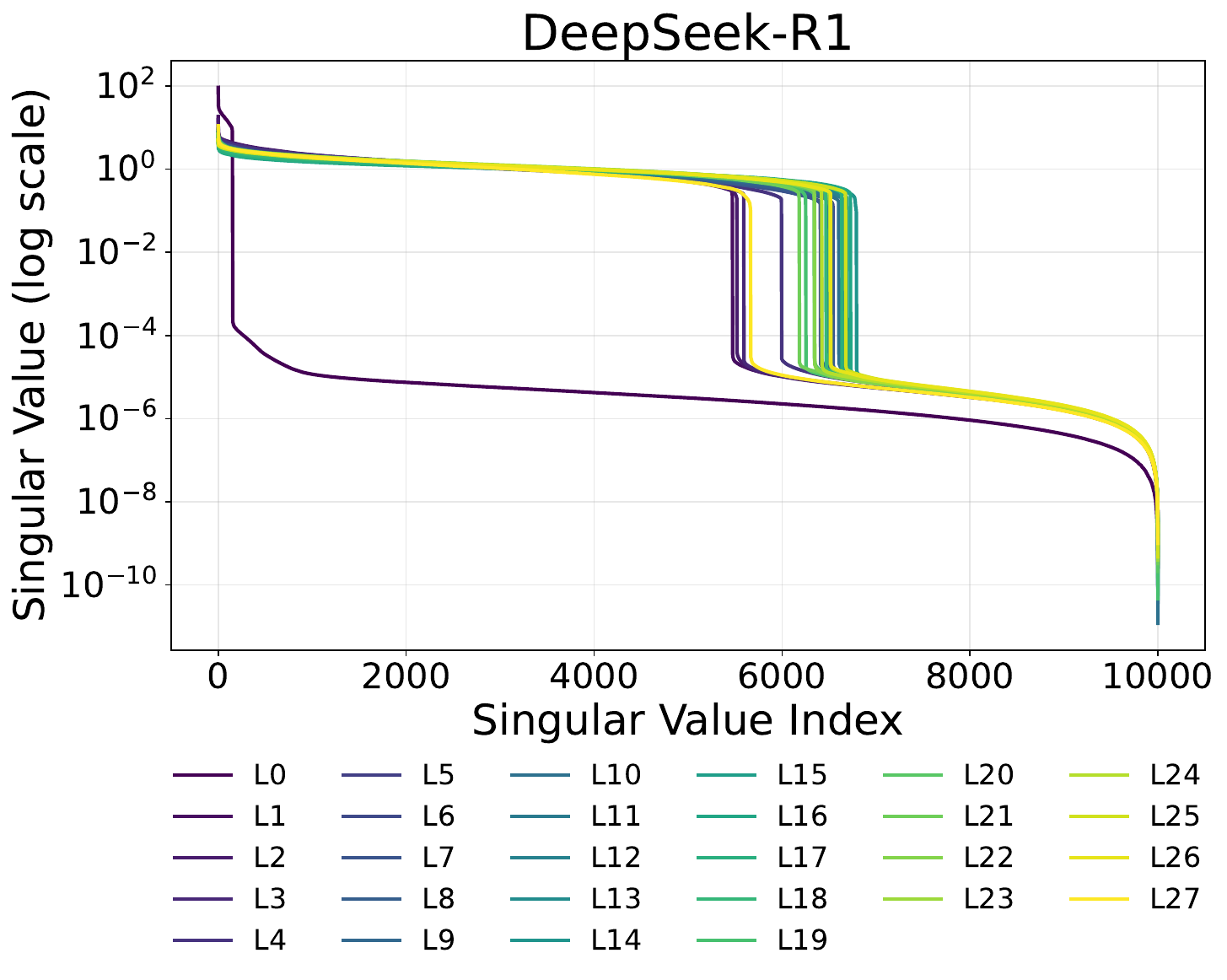}
\end{subfigure}
\caption{Singular value distributions in RFF space across layers. RFF spectral patterns reveal enhanced nonlinear characteristics compared to linear analysis: all models maintain higher effective ranks throughout layers, with GPT-2 showing reduced compression, RoBERTa exhibiting more uniform distributions, and both Llama and DeepSeek demonstrating sustained high-dimensional representations in kernel space.}
\label{fig:rff_svd_dist}
\end{figure*}

To understand nonlinear transformation properties, we analyze singular value distributions in RFF space using Random Fourier Features with RBF kernels. \Cref{fig:rff_svd_dist} shows spectral evolution across layers for all four architectures, computed from RFF-transformed representations. Comparing with linear spectral analysis (\Cref{fig:svd_dist}) reveals fundamental differences in nonlinear transformation characteristics: (1) RFF spectra maintain consistently higher effective ranks across all layers—while linear analysis showed dramatic compression in middle layers for decoder models, RFF distributions remain relatively stable, demonstrating that nonlinear transformations preserve more dimensional complexity than linear projections suggest; (2) Spectral decay patterns in RFF space are more gradual across all architectures, with GPT-2's characteristic middle-layer compression significantly reduced and RoBERTa maintaining even more uniform distributions, indicating kernel methods capture richer transformation structures; (3) The preservation of spectral mass at higher singular value indices in RFF space explains the consistently lower residual norms observed in our main results—RFF approximations utilize more principal components effectively, leading to better reconstruction quality; (4) Architectural differences persist but are less pronounced in kernel space, suggesting that while fundamental processing strategies differ between encoder and decoder models, nonlinear transformations add substantial complexity beyond what linear analysis reveals. These findings validate that RFF analysis provides complementary insights by capturing nonlinear transformation properties that enrich our understanding of transformer layer functions beyond linear approximations.

\subsection{Effective Rank Threshold Sensitivity Analysis}

\begin{figure*}[ht]
\centering
\begin{subfigure}[b]{0.495\textwidth}
    \centering
    \includegraphics[width=\textwidth]{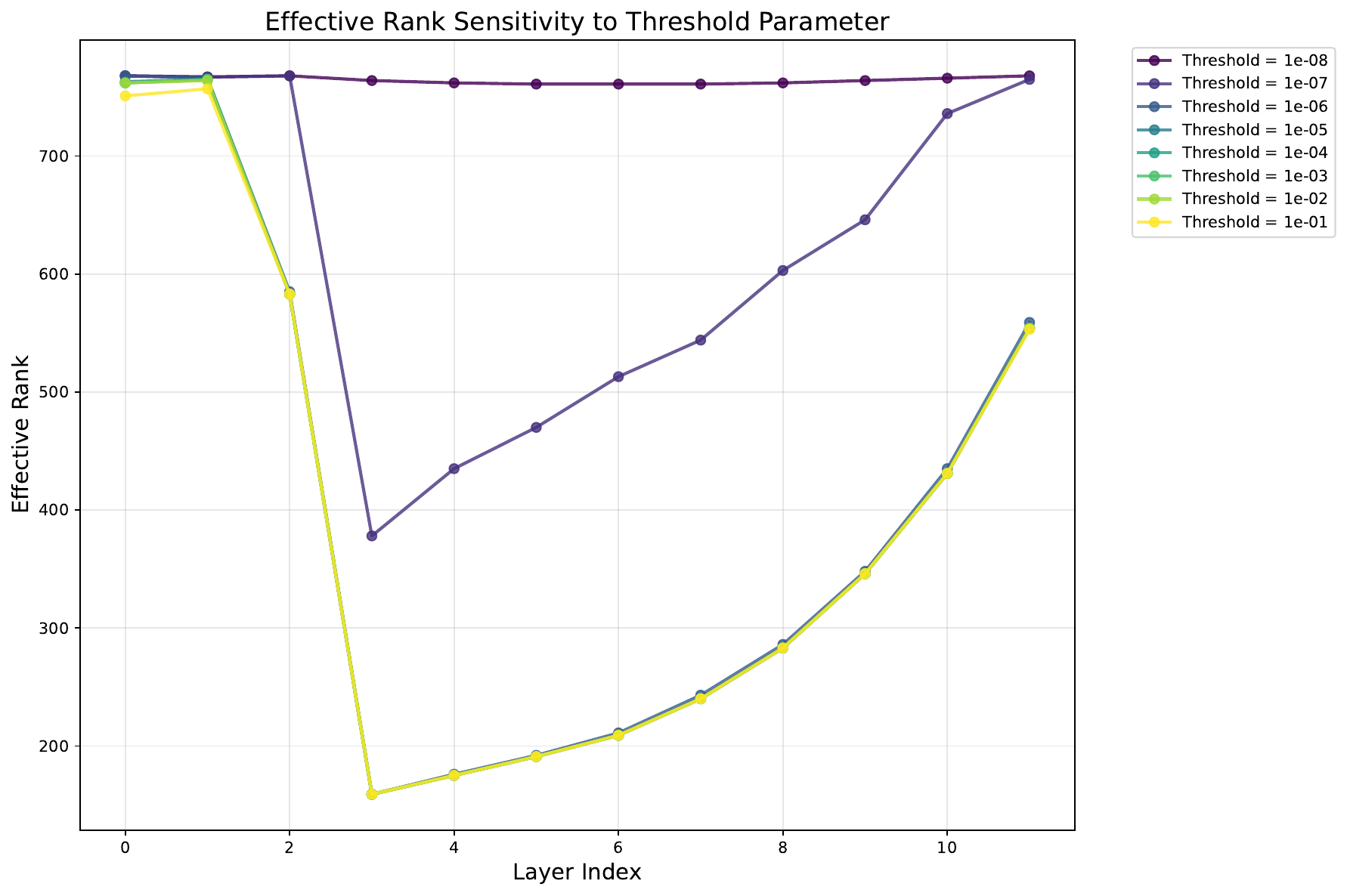}
    \caption{Effective rank vs layer index for different thresholds}
\end{subfigure}
\hfill
\begin{subfigure}[b]{0.495\textwidth}
    \centering
    \includegraphics[width=\textwidth]{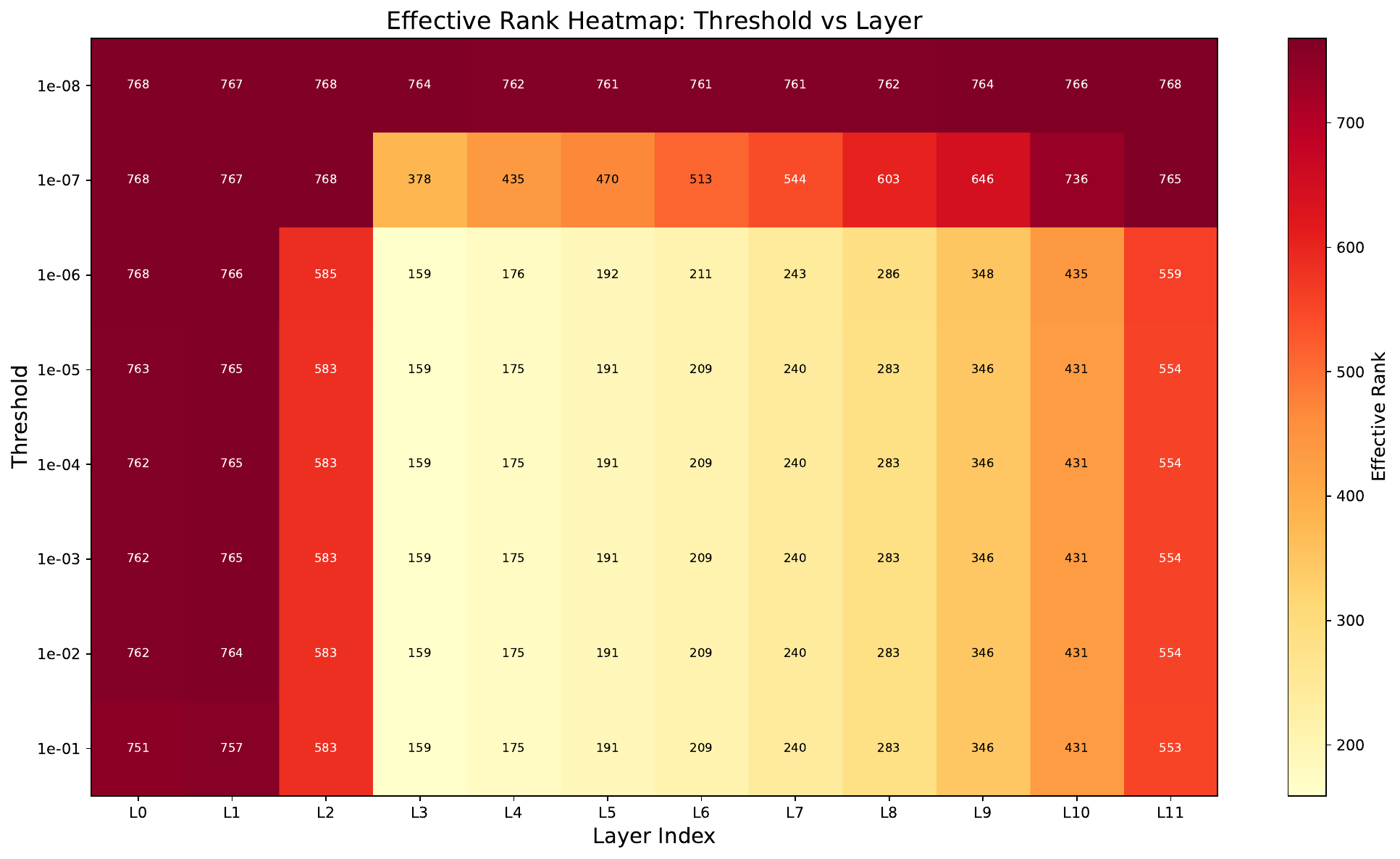}
    \caption{Threshold sensitivity heatmap}
\end{subfigure}
\caption{Effective rank sensitivity to threshold parameter across GPT-2 layers. (a) Shows how effective rank varies with threshold values from $10^{-8}$ to $10^{-1}$, revealing that early layers maintain stable estimates while middle layers exhibit high sensitivity. (b) Heatmap visualization demonstrates layer-specific sensitivity patterns, with layers 3-10 showing the highest variation (CV > 20\%) while layers 0-1 remain stable (CV < 5\%).}
\label{fig:threshold_sensitivity}
\end{figure*}

The effective rank metric depends on a threshold parameter $\epsilon$ that determines which singular values are considered significant. To assess the robustness of our findings, we conduct comprehensive sensitivity analysis across eight threshold values spanning five orders of magnitude: $\{10^{-8}, 10^{-7}, 10^{-6}, 10^{-5}, 10^{-4}, 10^{-3}, 10^{-2}, 10^{-1}\}$. \Cref{fig:threshold_sensitivity} presents the results from GPT-2 analysis with 500 samples per configuration.

Our analysis reveals three distinct sensitivity patterns across network depth: (1) \textbf{Stable layers} (0-1): Early layers exhibit remarkable stability with coefficient of variation (CV) below 5\%, maintaining effective ranks around 763-764 regardless of threshold choice. This stability indicates that early transformations preserve most singular values well above typical threshold ranges, suggesting robust high-dimensional processing; (2) \textbf{Sensitive layers} (3-10): Middle layers show high sensitivity with CV exceeding 20\%, where effective rank estimates vary dramatically—layer 3 ranges from 64 (at $\epsilon=10^{-1}$) to 507 (at $\epsilon=10^{-8}$). This sensitivity reflects the compression bottleneck identified in our main analysis, where many singular values cluster near threshold boundaries; (3) \textbf{Moderate layers} (2, 11): Transition layers exhibit intermediate sensitivity (CV 12-15\%), bridging stable and sensitive regions.

The threshold sensitivity patterns provide additional validation for our architectural findings. The high sensitivity in middle layers corresponds precisely to the compression phase identified in decoder architectures, where dimensional reduction creates singular values distributed across multiple magnitude scales. Conversely, the stability of early and late layers confirms that high-rank and expansion phases produce well-separated singular values robust to threshold variation. Based on stability analysis across all tested values, we recommend threshold range $[10^{-7}, 10^{-2}]$ for practical applications, excluding extreme values that may capture numerical noise (below $10^{-7}$) or overly aggressive filtering (above $10^{-2}$). Our default choice of $\epsilon = 10^{-5}$ lies centrally within this stable range and produces average effective rank of 391.6 across layers, consistent with the compression-expansion patterns reported in our main results. This sensitivity analysis strengthens our conclusions by demonstrating that the identified architectural patterns persist across reasonable threshold choices, though practitioners should consider layer-specific sensitivity when interpreting middle-layer measurements.

\subsection{RFF Dimensions Parameter Sensitivity}

\begin{figure*}[ht]
\centering
\includegraphics[width=0.9\textwidth]{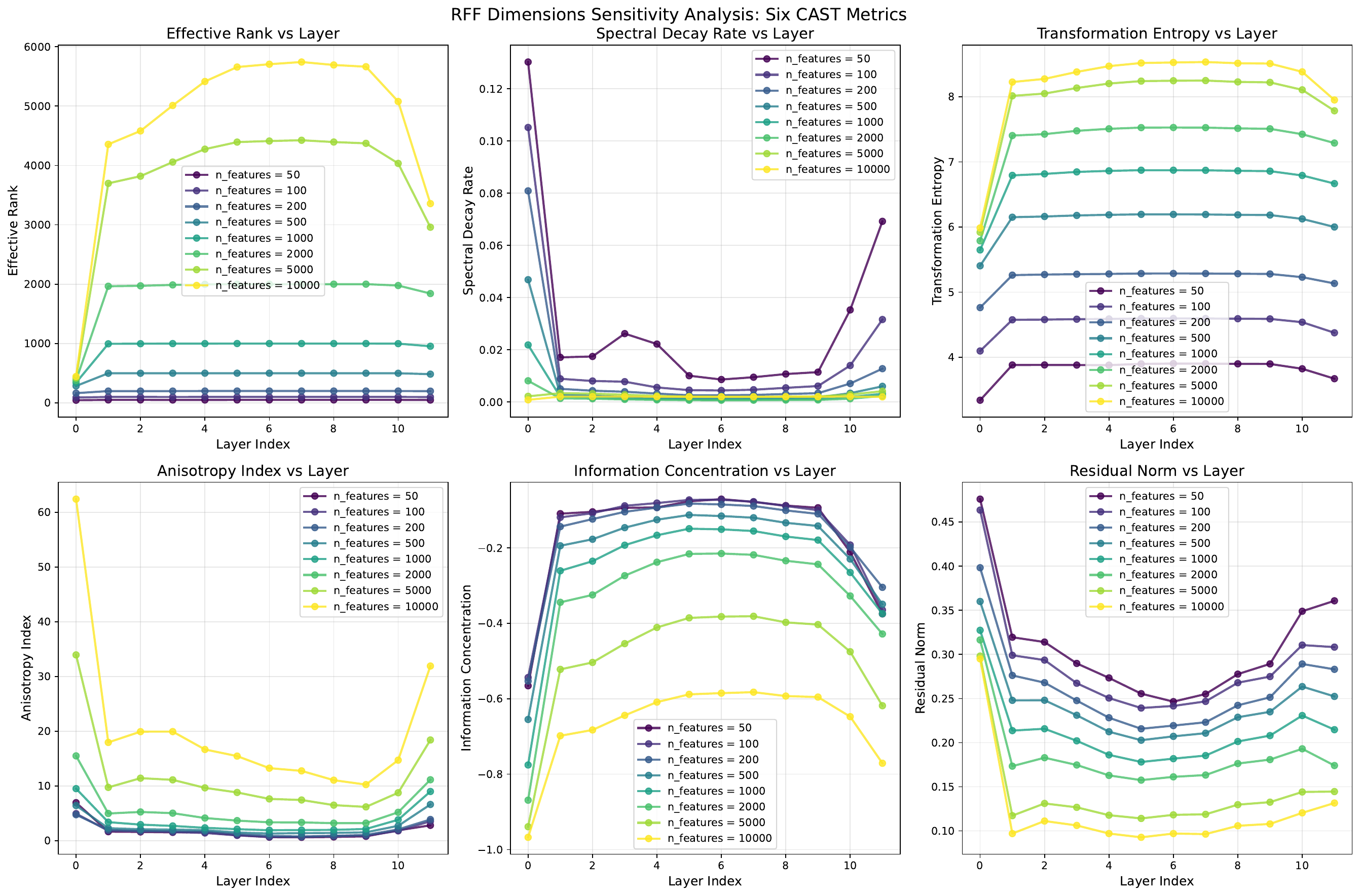}
\caption{RFF dimensions sensitivity analysis showing how all six CAST metrics vary with the number of Random Fourier Features. The plots demonstrate systematic changes in metric behavior as n\_features increases from 50 to 10,000, revealing convergence patterns and computational trade-offs. Each metric shows distinct scaling behavior: effective rank grows linearly, spectral decay rate decreases exponentially, transformation entropy increases logarithmically, anisotropy index shows power-law growth, information concentration becomes increasingly negative, and residual norm decreases asymptotically.}
\label{fig:rff_dimensions}
\end{figure*}

Random Fourier Features (RFF) approximation quality depends critically on the number of features $d$ used to approximate the kernel. To understand this dependency and establish practical guidelines, we conduct comprehensive sensitivity analysis across eight n\_features values spanning four orders of magnitude: $\{50, 100, 200, 500, 1000, 2000, 5000, 10000\}$. \Cref{fig:rff_dimensions} presents the systematic variation of all six CAST metrics across GPT-2 layers for different approximation dimensions.

Our analysis reveals distinct scaling behaviors for each metric: (1) \textbf{Effective Rank}: Shows near-perfect linear scaling with n\_features, growing from 49.3 (n=50) to 4724.8 (n=10000). This linear relationship confirms that RFF preserves the dimensional structure of the kernel space, with effective rank bounded by the approximation dimension; (2) \textbf{Spectral Decay Rate}: Exhibits exponential decrease from 0.0307 to 0.0019, indicating that higher-dimensional RFF approximations capture more fine-grained spectral structure with slower decay patterns; (3) \textbf{Transformation Entropy}: Increases logarithmically from 3.82 to 8.19, reflecting the enhanced capacity to capture transformation complexity in higher-dimensional feature spaces; (4) \textbf{Anisotropy Index}: Demonstrates power-law growth from 1.80 to 20.52, suggesting that high-dimensional RFF spaces reveal increasingly anisotropic transformation patterns previously obscured in lower dimensions; (5) \textbf{Information Concentration}: Becomes progressively more negative (-0.16 to -0.66), indicating better information distribution across the expanded feature space; (6) \textbf{Residual Norm}: Shows asymptotic decrease from 0.31 to 0.12, confirming improved approximation quality with diminishing returns at higher dimensions.

The scaling analysis provides crucial insights for practical RFF implementation. While approximation quality improves monotonically with n\_features, computational cost scales linearly, creating important trade-offs. Based on convergence analysis, we observe that: (1) n\_features = 200 provides acceptable approximation quality for rapid prototyping; (2) n\_features = 1000 offers good balance between accuracy and computational efficiency for most research applications; (3) n\_features $\ge$ 5000 yields high-fidelity approximations suitable for detailed analysis, though with significantly increased computational burden. The systematic metric scaling also validates our kernel approximation approach—the predictable mathematical relationships between n\_features and metric values demonstrate that RFF reliably captures the underlying kernel structure across different approximation qualities. These findings enable practitioners to select appropriate n\_features values based on their specific accuracy requirements and computational constraints, with clear understanding of the resulting metric behavior changes.

%% file: arxiv.bbl
\begin{thebibliography}{49}
\providecommand{\natexlab}[1]{#1}
\providecommand{\url}[1]{\texttt{#1}}
\expandafter\ifx\csname urlstyle\endcsname\relax
  \providecommand{\doi}[1]{doi: #1}\else
  \providecommand{\doi}{doi: \begingroup \urlstyle{rm}\Url}\fi

\bibitem[Ansuini et~al.(2019)Ansuini, Laio, Macke, and Zoccolan]{ansuini2019intrinsic}
Alessio Ansuini, Alessandro Laio, Jakob~H Macke, and Davide Zoccolan.
\newblock Intrinsic dimension of data representations in deep neural networks.
\newblock In \emph{Advances in Neural Information Processing Systems}, pages 6111--6122, 2019.

\bibitem[Belinkov(2022)]{belinkov2022probing}
Yonatan Belinkov.
\newblock Probing classifiers: Promises, shortcomings, and advances.
\newblock \emph{Computational Linguistics}, 48\penalty0 (1):\penalty0 207--219, 2022.

\bibitem[Belinkov and Glass(2019)]{belinkov2019analysis}
Yonatan Belinkov and James Glass.
\newblock Analysis methods in neural language processing: A survey.
\newblock \emph{Transactions of the Association for Computational Linguistics}, 7:\penalty0 49--72, 2019.

\bibitem[Belrose et~al.(2023)Belrose, Furman, Smith, Halawi, Ostrovsky, McKinney, Biderman, and Steinhardt]{belrose2023eliciting}
Nora Belrose, Zach Furman, Logan Smith, Danny Halawi, Igor Ostrovsky, Lev McKinney, Stella Biderman, and Jacob Steinhardt.
\newblock Eliciting latent predictions from transformers with the tuned lens.
\newblock \emph{arXiv preprint arXiv:2303.08112}, 2023.

\bibitem[Bengio et~al.(2013)Bengio, Courville, and Vincent]{bengio2013representation}
Yoshua Bengio, Aaron Courville, and Pascal Vincent.
\newblock Representation learning: A review and new perspectives.
\newblock \emph{IEEE transactions on pattern analysis and machine intelligence}, 35\penalty0 (8):\penalty0 1798--1828, 2013.

\bibitem[Bloom and Sharma(2022)]{bloom2023singular}
Jake Bloom and Siddharth Sharma.
\newblock The singular value decompositions of transformer weight matrices are highly interpretable.
\newblock \emph{AI Alignment Forum}, 2022.

\bibitem[Bricken et~al.(2023)Bricken, Templeton, Batson, Chen, Jermyn, Conerly, Turner, Anil, Denison, Askell, Lasenby, Wu, Kravec, Schiefer, Maxwell, Joseph, Tamkin, Nguyen, McLean, Burke, Hume, Carter, Henighan, and Olah]{bricken2023monosemanticity}
Trenton Bricken, Adly Templeton, Joshua Batson, Brian Chen, Adam Jermyn, Tom Conerly, Nicholas~L Turner, Cem Anil, Carson Denison, Amanda Askell, Robert Lasenby, Yifan Wu, Shauna Kravec, Nicholas Schiefer, Tim Maxwell, Nicholas Joseph, Alex Tamkin, Karina Nguyen, Brayden McLean, Josiah~E Burke, Tristan Hume, Shan Carter, Tom Henighan, and Chris Olah.
\newblock Towards monosemanticity: Decomposing language models with dictionary learning.
\newblock \emph{Transformer Circuits Thread}, 2023.

\bibitem[Chen et~al.(2022)Chen, Xie, Song, Chen, and Tang]{chen2022joint}
Yingxi Chen, Yunhe Xie, Linhai Song, Fan Chen, and Tie Tang.
\newblock Joint matrix decomposition for deep convolutional neural networks compression.
\newblock \emph{Neurocomputing}, 516:\penalty0 11--24, 2022.

\bibitem[Conmy et~al.(2023)Conmy, Mavor-Parker, Lynch, Heimersheim, and Garriga-Alonso]{conmy2023automated}
Arthur Conmy, Augustine~N Mavor-Parker, Aengus Lynch, Stefan Heimersheim, and Adri{\`a} Garriga-Alonso.
\newblock Towards automated circuit discovery for mechanistic interpretability.
\newblock In \emph{Advances in Neural Information Processing Systems}, 2023.

\bibitem[Cunningham et~al.(2023)Cunningham, Ewart, Riggs, Huben, and Sharkey]{cunningham2023sparse}
Hoagy Cunningham, Aidan Ewart, Logan Riggs, Robert Huben, and Lee Sharkey.
\newblock Sparse autoencoders find highly interpretable features in language models.
\newblock \emph{arXiv preprint arXiv:2309.08600}, 2023.

\bibitem[{DeepSeek-AI Team}(2025)]{deepseek2025}
{DeepSeek-AI Team}.
\newblock Deepseek-r1: Incentivizing reasoning capability in llms via reinforcement learning.
\newblock \emph{arXiv preprint arXiv:2501.12948}, 2025.

\bibitem[Denton et~al.(2014)Denton, Zaremba, Bruna, LeCun, and Fergus]{denton2014exploiting}
Emily~L Denton, Wojciech Zaremba, Joan Bruna, Yann LeCun, and Rob Fergus.
\newblock Exploiting linear structure within convolutional networks for efficient evaluation.
\newblock In \emph{Advances in neural information processing systems}, pages 1269--1277, 2014.

\bibitem[Drineas and Mahoney(2016)]{drineas2016fast}
Petros Drineas and Michael~W Mahoney.
\newblock Fast randomized algorithms for the approximation of matrices.
\newblock \emph{Communications of the ACM}, 59\penalty0 (6):\penalty0 84--93, 2016.

\bibitem[Elazar et~al.(2021)Elazar, Ravfogel, Jacovi, and Goldberg]{elazar2021amnesic}
Yanai Elazar, Shauli Ravfogel, Alon Jacovi, and Yoav Goldberg.
\newblock Amnesic probing: Behavioral explanation with amnesic counterfactuals.
\newblock \emph{Transactions of the Association for Computational Linguistics}, 9:\penalty0 160--175, 2021.

\bibitem[Elhage et~al.(2021)Elhage, Nanda, Olsson, Henighan, Joseph, Mann, Askell, Bai, Chen, Conerly, DasSarma, Drain, Ganguli, Hatfield-Dodds, Hernandez, Jones, Kernion, Lovitt, Ndousse, Amodei, Brown, Clark, Kaplan, McCandlish, and Olah]{elhage2021mathematical}
Nelson Elhage, Neel Nanda, Catherine Olsson, Tom Henighan, Nicholas Joseph, Ben Mann, Amanda Askell, Yuntao Bai, Anna Chen, Tom Conerly, Nova DasSarma, Dawn Drain, Deep Ganguli, Zac Hatfield-Dodds, Danny Hernandez, Andy Jones, Jackson Kernion, Liane Lovitt, Kamal Ndousse, Dario Amodei, Tom Brown, Jack Clark, Jared Kaplan, Sam McCandlish, and Chris Olah.
\newblock A mathematical framework for transformer circuits.
\newblock \emph{Transformer Circuits Thread}, 2021.

\bibitem[Ethayarajh(2019)]{ethayarajh2019how}
Kawin Ethayarajh.
\newblock How contextual are contextualized word representations? comparing the geometry of bert, elmo, and gpt-2 embeddings.
\newblock In \emph{Proceedings of the 2019 Conference on Empirical Methods in Natural Language Processing and the 9th International Joint Conference on Natural Language Processing (EMNLP-IJCNLP)}, pages 55--65, 2019.

\bibitem[Ghandeharioun et~al.(2024)Ghandeharioun, Caciularu, Pearson, Dixon, and Gehrmann]{ghandeharioun2024patchscopes}
Asma Ghandeharioun, Avi Caciularu, Adam Pearson, Lucas Dixon, and Sebastian Gehrmann.
\newblock Patchscopes: A unifying framework for inspecting hidden representations of language models.
\newblock In \emph{Proceedings of the 41st International Conference on Machine Learning}, 2024.

\bibitem[Golub and Van~Loan(2013)]{golub2013matrix}
Gene~H Golub and Charles~F Van~Loan.
\newblock \emph{Matrix Computations}.
\newblock Johns Hopkins University Press, 4th edition, 2013.

\bibitem[Hewitt and Manning(2019)]{hewitt2019structural}
John Hewitt and Christopher~D Manning.
\newblock A structural probe for finding syntax in word representations.
\newblock In \emph{Proceedings of NAACL-HLT}, pages 4129--4138, 2019.

\bibitem[Hoerl and Kennard(1970)]{hoerl1970ridge}
Arthur~E Hoerl and Robert~W Kennard.
\newblock Ridge regression: Biased estimation for nonorthogonal problems.
\newblock \emph{Technometrics}, 12\penalty0 (1):\penalty0 55--67, 1970.

\bibitem[Jiang et~al.(2020)Jiang, Xu, Araki, and Neubig]{jiang2020can}
Zhengbao Jiang, Frank~F Xu, Jun Araki, and Graham Neubig.
\newblock Inserting information bottlenecks for attribution in transformers.
\newblock In \emph{Findings of the Association for Computational Linguistics: EMNLP 2020}, pages 4246--4251, 2020.

\bibitem[Johansson et~al.(2022)Johansson, Enroth, and Mogren]{johansson2022exact}
Anton Johansson, Hannah Enroth, and Olof Mogren.
\newblock Exact spectral norm regularization for neural networks.
\newblock \emph{arXiv preprint arXiv:2206.13581}, 2022.

\bibitem[Kornblith et~al.(2019)Kornblith, Norouzi, Lee, and Hinton]{kornblith2019similarity}
Simon Kornblith, Mohammad Norouzi, Honglak Lee, and Geoffrey Hinton.
\newblock Similarity of neural network representations revisited.
\newblock In \emph{International Conference on Machine Learning}, pages 3519--3529, 2019.

\bibitem[Kovaleva et~al.(2019)Kovaleva, Romanov, Rogers, and Rumshisky]{kovaleva2019revealing}
Olga Kovaleva, Alexey Romanov, Anna Rogers, and Anna Rumshisky.
\newblock Revealing the dark secrets of bert.
\newblock \emph{arXiv preprint arXiv:1908.08593}, 2019.

\bibitem[Li et~al.(2025)Li, Xia, Zhang, Hui, Kong, Zhang, and Yang]{adasvd2025}
Zhiteng Li, Mingyuan Xia, Jingyuan Zhang, Zheng Hui, Linghe Kong, Yulun Zhang, and Xiaokang Yang.
\newblock Adasvd: Adaptive singular value decomposition for large language models.
\newblock \emph{arXiv preprint arXiv:2502.01403}, 2025.

\bibitem[Liu et~al.(2019)Liu, Ott, Goyal, Du, Joshi, Chen, Levy, Lewis, Zettlemoyer, and Stoyanov]{liu2019roberta}
Yinhan Liu, Myle Ott, Naman Goyal, Jingfei Du, Mandar Joshi, Danqi Chen, Omer Levy, Mike Lewis, Luke Zettlemoyer, and Veselin Stoyanov.
\newblock Roberta: A robustly optimized bert pretraining approach.
\newblock \emph{arXiv preprint arXiv:1907.11692}, 2019.

\bibitem[Machina and Mercer(2024)]{machina2024anisotropy}
Aaron Machina and Robert~E Mercer.
\newblock Anisotropy is not inherent to transformers.
\newblock In \emph{Proceedings of the 2024 Conference of the North American Chapter of the Association for Computational Linguistics: Human Language Technologies (Volume 1: Long Papers)}, pages 5364--5375, 2024.

\bibitem[Merity et~al.(2017)Merity, Xiong, Bradbury, and Socher]{merity2017pointer}
Stephen Merity, Caiming Xiong, James Bradbury, and Richard Socher.
\newblock Pointer sentinel mixture models.
\newblock \emph{arXiv preprint arXiv:1609.07843}, 2017.

\bibitem[{nostalgebraist}(2020)]{nostalgebraist2020logit}
{nostalgebraist}.
\newblock interpreting gpt: the logit lens.
\newblock \emph{LessWrong}, 2020.
\newblock URL \url{https://www.lesswrong.com/posts/AcKRB8wDpdaN6v6ru/interpreting-gpt-the-logit-lens}.

\bibitem[Olah(2023)]{olah2023interpretability}
Chris Olah.
\newblock Interpretability dreams.
\newblock \emph{Transformer Circuits Thread}, 2023.

\bibitem[Olah et~al.(2020)Olah, Cammarata, Schubert, Goh, Petrov, and Carter]{olah2020zoom}
Chris Olah, Nick Cammarata, Ludwig Schubert, Gabriel Goh, Michael Petrov, and Shan Carter.
\newblock Zoom in: An introduction to circuits.
\newblock \emph{Distill}, 5\penalty0 (3):\penalty0 e00024--001, 2020.

\bibitem[Radford et~al.(2019)Radford, Wu, Child, Luan, Amodei, and Sutskever]{radford2019language}
Alec Radford, Jeffrey Wu, Rewon Child, David Luan, Dario Amodei, and Ilya Sutskever.
\newblock Language models are unsupervised multitask learners.
\newblock \emph{OpenAI blog}, 1\penalty0 (8):\penalty0 9, 2019.

\bibitem[Rahimi and Recht(2007)]{rahimi2007random}
Ali Rahimi and Benjamin Recht.
\newblock Random features for large-scale kernel machines.
\newblock In \emph{Advances in neural information processing systems}, pages 1177--1184, 2007.

\bibitem[Ravichander et~al.(2021)Ravichander, Belinkov, and Hovy]{ravichander2021probing}
Abhilasha Ravichander, Yonatan Belinkov, and Eduard Hovy.
\newblock Probing the probing paradigm: Does probing accuracy entail task relevance?
\newblock In \emph{Proceedings of the 16th Conference of the European Chapter of the Association for Computational Linguistics: Main Volume}, pages 3363--3377, 2021.

\bibitem[Razzhigaev et~al.(2024)Razzhigaev, Kazmina, Volkova, Cherkasova, Kirilov, Tikhonov, Panchenko, Oseledets, and Burnaev]{razzhigaev2024shape}
Anton Razzhigaev, Matvey Kazmina, Alena Volkova, Natalia Cherkasova, Konstantin Kirilov, Mikhail Tikhonov, Artem Panchenko, Ivan Oseledets, and Evgeny Burnaev.
\newblock The shape of learning: Anisotropy and intrinsic dimensions in transformer-based models.
\newblock In \emph{Findings of the Association for Computational Linguistics: EACL 2024}, pages 961--980, 2024.

\bibitem[Rogers et~al.(2020)Rogers, Kovaleva, and Rumshisky]{rogers2020primer}
Anna Rogers, Olga Kovaleva, and Anna Rumshisky.
\newblock A primer in bertology: What we know about how bert works.
\newblock \emph{Transactions of the Association for Computational Linguistics}, 8:\penalty0 842--866, 2020.

\bibitem[Roy and Vetterli(2007)]{roy2007effective}
Olivier Roy and Martin Vetterli.
\newblock Effective rank: A measure of effective dimensionality.
\newblock \emph{15th European Signal Processing Conference}, pages 606--610, 2007.

\bibitem[Schwartz-Ziv and Tishby(2017)]{schwartz2017opening}
Ravid Schwartz-Ziv and Naftali Tishby.
\newblock Opening the black box of deep neural networks via information.
\newblock \emph{arXiv preprint arXiv:1703.00810}, 2017.

\bibitem[Templeton et~al.(2024)Templeton, Conerly, Marcus, Lindsey, Bricken, Chen, Pearce, Citro, Ameisen, Jones, Cunningham, Turner, McDougall, MacDiarmid, Tamkin, Durmus, Hume, Mosconi, Freeman, Sumers, Rees, Batson, Jermyn, Carter, Olah, and Henighan]{templeton2024scaling}
Adly Templeton, Tom Conerly, Jonathan Marcus, Jack Lindsey, Trenton Bricken, Brian Chen, Adam Pearce, Craig Citro, Emmanuel Ameisen, Andy Jones, Hoagy Cunningham, Nicholas~L Turner, Callum McDougall, Monte MacDiarmid, Alex Tamkin, Esin Durmus, Tristan Hume, Francesco Mosconi, C.~Daniel Freeman, Theodore~R Sumers, Edward Rees, Joshua Batson, Adam Jermyn, Shan Carter, Chris Olah, and Tom Henighan.
\newblock Scaling monosemanticity: Extracting interpretable features from claude 3 sonnet.
\newblock \emph{Transformer Circuits Thread}, 2024.

\bibitem[Tenney et~al.(2019)Tenney, Das, and Pavlick]{tenney2019bert}
Ian Tenney, Dipanjan Das, and Ellie Pavlick.
\newblock Bert rediscovers the classical nlp pipeline.
\newblock \emph{arXiv preprint arXiv:1905.05950}, 2019.

\bibitem[Tishby and Zaslavsky(2015)]{tishby2015deep}
Naftali Tishby and Noga Zaslavsky.
\newblock Deep learning and the information bottleneck principle.
\newblock \emph{arXiv preprint arXiv:1503.02406}, 2015.

\bibitem[Touvron et~al.(2023)Touvron, Martin, Stone, Albert, Almahairi, Babaei, Bashlykov, Batra, Bhargava, Bhosale, Bikel, Blecher, Ferrer, Chen, Cucurull, Esiobu, Fernandes, Fu, Fu, Fuller, Gao, Goswami, Goyal, Hartshorn, Hosseini, Hou, Inan, Kardas, Kerkez, Khabsa, Kloumann, Korenev, Koura, Lachaux, Lavril, Lee, Liskovich, Lu, Mao, Martinet, Mihaylov, Mishra, Molybog, Nie, Poulton, Reizenstein, Rungta, Saladi, Schelten, Silva, Smith, Subramanian, Tan, Tang, Taylor, Williams, Kuan, Xu, Yan, Zarov, Zhang, Fan, Kambadur, Narang, Rodriguez, Stojnic, Edunov, and Scialom]{touvron2023llama}
Hugo Touvron, Louis Martin, Kevin Stone, Peter Albert, Amjad Almahairi, Yasmine Babaei, Nikolay Bashlykov, Soumya Batra, Prajjwal Bhargava, Shruti Bhosale, Dan Bikel, Lukas Blecher, Cristian~Canton Ferrer, Moya Chen, Guillem Cucurull, David Esiobu, Jude Fernandes, Jeremy Fu, Wenyin Fu, Brian Fuller, Cynthia Gao, Vedanuj Goswami, Naman Goyal, Anthony Hartshorn, Saghar Hosseini, Rui Hou, Hakan Inan, Marcin Kardas, Viktor Kerkez, Madian Khabsa, Isabel Kloumann, Artem Korenev, Punit~Singh Koura, Marie-Anne Lachaux, Thibaut Lavril, Jenya Lee, Diana Liskovich, Yinghai Lu, Yuning Mao, Xavier Martinet, Todor Mihaylov, Pushkar Mishra, Igor Molybog, Yixin Nie, Andrew Poulton, Jeremy Reizenstein, Rashi Rungta, Kalyan Saladi, Alan Schelten, Ruan Silva, Eric~Michael Smith, Ranjan Subramanian, Xiaoqing~Ellen Tan, Binh Tang, Ross Taylor, Adina Williams, Jian~Xiang Kuan, Puxin Xu, Zheng Yan, Iliyan Zarov, Yuchen Zhang, Angela Fan, Melanie Kambadur, Sharan Narang, Aurelien Rodriguez, Robert Stojnic, Sergey Edunov, and Thomas Scialom.
\newblock Llama 2: Open foundation and fine-tuned chat models.
\newblock \emph{arXiv preprint arXiv:2307.09288}, 2023.

\bibitem[Ubaru et~al.(2017)Ubaru, Chen, and Saad]{ubaru2016fast}
Shashanka Ubaru, Jie Chen, and Yousef Saad.
\newblock Fast estimation of tr(f(a)) via stochastic lanczos quadrature.
\newblock \emph{SIAM Journal on Matrix Analysis and Applications}, 38\penalty0 (4):\penalty0 1075--1099, 2017.

\bibitem[Udell and Townsend(2019)]{udell2019matrix}
Madeleine Udell and Alex Townsend.
\newblock Why are big data matrices approximately low rank?
\newblock \emph{SIAM Journal on Mathematics of Data Science}, 1\penalty0 (1):\penalty0 144--160, 2019.

\bibitem[Vig(2019)]{vig2019multiscale}
Jesse Vig.
\newblock A multiscale visualization of attention in the transformer model.
\newblock In \emph{Proceedings of the 57th Annual Meeting of the Association for Computational Linguistics: System Demonstrations}, pages 37--42, 2019.

\bibitem[Vig and Belinkov(2019)]{vig2019analyzing}
Jesse Vig and Yonatan Belinkov.
\newblock Analyzing the structure of attention in a transformer language model.
\newblock In \emph{Proceedings of the 2019 ACL Workshop BlackboxNLP: Analyzing and Interpreting Neural Networks for NLP}, pages 63--76, 2019.

\bibitem[Voita et~al.(2019)Voita, Sennrich, and Titov]{voita2019bottom}
Elena Voita, Rico Sennrich, and Ivan Titov.
\newblock The bottom-up evolution of representations in the transformer: A study with machine translation and language modeling objectives.
\newblock \emph{arXiv preprint arXiv:1909.01380}, 2019.

\bibitem[Zhou and Srikumar(2021)]{zhou2021directprobe}
Yichu Zhou and Vivek Srikumar.
\newblock Directprobe: Studying representations without classifiers.
\newblock \emph{arXiv preprint arXiv:2104.03514}, 2021.

\bibitem[Zou and Hastie(2005)]{zou2005regularization}
Hui Zou and Trevor Hastie.
\newblock Regularization and variable selection via the elastic net.
\newblock \emph{Journal of the Royal Statistical Society Series B: Statistical Methodology}, 67\penalty0 (2):\penalty0 301--320, 2005.

\end{thebibliography}
